\documentclass{article}

\usepackage{microtype}
\usepackage{graphicx}
\usepackage{subfigure}
\usepackage{booktabs} 

\usepackage{hyperref}

\usepackage[accepted]{icml2023}

\usepackage{amsmath}
\usepackage{amssymb}
\usepackage{mathtools}
\usepackage{amsthm}
\usepackage{bbm}

\usepackage[capitalize,noabbrev]{cleveref}

\usepackage{kotex}
\usepackage{multirow}
\usepackage{marvosym}
\usepackage{makecell}
\usepackage{comment}
\usepackage{enumitem}
\usepackage{enumerate}

\theoremstyle{plain}

\theoremstyle{definition}

\theoremstyle{remark}

\usepackage[textsize=tiny]{todonotes}

\icmltitlerunning{One-shot Imitation in a Non-Stationary Environment via Multi-Modal Skill}

\begin{document}

\twocolumn[
\icmltitle{One-shot Imitation in a Non-Stationary Environment via Multi-Modal Skill}





\icmlsetsymbol{equal}{*}

\begin{icmlauthorlist}
\icmlauthor{Sangwoo Shin}{skku}
\icmlauthor{Daehee Lee}{skku}
\icmlauthor{Minjong Yoo}{skku}
\icmlauthor{Woo Kyung Kim}{skku}
\icmlauthor{Honguk Woo}{skku}
\end{icmlauthorlist}

\icmlaffiliation{skku}{Department of Computer Science and Engineering, Sungkyunkwan University, Suwon, Republic of Korea}

\icmlcorrespondingauthor{Honguk Woo}{hwoo@skku.edu}

\icmlkeywords{Machine Learning, ICML}

\vskip 0.3in
]



\printAffiliationsAndNotice{}  

\begin{abstract}
%
One-shot imitation is to learn a new task from a single demonstration, yet it is a challenging problem to adopt it for complex tasks with the high domain diversity inherent in a non-stationary environment. 
To tackle the problem, we explore the compositionality of complex tasks, and present a novel skill-based imitation learning framework enabling one-shot imitation and zero-shot adaptation; from a single demonstration for a complex unseen task, a semantic skill sequence is inferred and then each skill in the sequence is converted into an action sequence optimized for 
 environmental hidden dynamics that can vary over time. 
Specifically, we leverage a vision-language model to learn a semantic skill set from offline video datasets, where each skill is represented on the vision-language embedding space, and adapt meta-learning with dynamics inference to enable zero-shot skill adaptation.   
We evaluate our framework with various one-shot imitation scenarios for extended multi-stage Meta-world tasks, showing its superiority in learning complex tasks, generalizing to dynamics changes, and extending to different demonstration conditions and modalities, compared to other baselines.
\end{abstract}

\section{Introduction}
In the context of learning control tasks for autonomous agents, one-shot imitation aims at learning such a task on a single demonstration presented~\cite{NIPS2017_ba386660oneshotimitation,  pmlr-v78-finn17aoneshotvisual, li2022metaimitation}.
One-shot imitation is particularly desirable, as it allows efficient use of expert demonstrations, which are often costly to collect for diverse tasks. However, it is considered challenging to achieve reliable one-shot imitation for complex tasks across various domains and environments~\cite{pertsch2022crossdomainstar, jang2022bcz, ichter2022dosaycan}.

In this paper, we investigate skill-based imitation learning techniques to conduct one-shot imitation for long-horizon tasks in a non-stationary environment, where underlying dynamics can be changed over time. To do so, we leverage the compositionality in control tasks and explore the semantic representation capability of a large scale vision-language pretrained model. 
Specifically, we develop a novel one-shot imitation learning framework using semantic skills, OnIS to enable one-shot imitation for new long-horizon tasks across different dynamics in a non-stationary environment.
The framework employs a vision-language pretrained model and contrastive learning on expert data to represent semantic skills and environment dynamics separately, and explores meta-learning to adapt semantic skills to unseen dynamics.
This establishes the capability of dynamics-aware skill transfer. 
In deployment, given a single video demonstration for a new task, the framework translates it to a sequence of semantic skills through a sequence encoder built on the vision-language model. 
It then combines the skills with inferred current dynamics to generate actions optimized for not only the demonstration but also the current dynamics.    

As such, through the decomposition of skills and dynamics, each skill can be well represented in the vision-language aligned embedding space without much deformation by time-varying environment dynamics. 
This decomposition structure with multi-modality renders OnIS unique to explore the compositionality of control tasks with skills, different from prior skill-based reinforcement learning (RL) approaches, e.g.,~\cite{spirl, simpl}.
%

The main contributions of our work are as follows. 
\begin{itemize}[noitemsep, topsep=0pt]
    \item We present the OnIS framework to enable one-shot imitation and zero-shot adaptation.    
    \item We develop a dynamics-aware skill transfer scheme to adapt skills in a non-stationary environment. 
    \item We create an expert dataset for long-horizon, multi-stage Meta-world tasks with diverse robotic manipulation skills, and make it publicly available for other research works.  
    \item We test the OnIS framework with several imitation learning use cases, and evaluate it in terms of one-shot imitation performance, generalization ability to different environment conditions, and extensibility to demonstrations in different modalities. 
\end{itemize}

\section{Problem Formulation} \label{sec:problem}
%
In vision-based tasks, one-shot imitation aims at learning a task upon a single expert demonstration presented in a video clip. 
Among several related problems in one-shot imitation learning approaches, our work considers a challenging yet practical problem setting, inherent in tackling complex tasks via imitation in a non-stationary environment. 

Unlike recent one-shot imitation approaches, our problem setting requires the followings to be addressed in a unified framework. 
(1) A policy should not only be learned from a single expert demonstration, but also adapt to unseen dynamics conditions inherent in a non-stationary environment,
(2) an expert demonstration specifies a long-horizon multi-stage task, and
(3) an expert demonstration can be presented either in video or language.  
Here, we formally describe our problem, ``one-shot imitation and zero-shot adaptation'' in a non-stationary environment. 
\begin{figure}[t]
    \centering
    \includegraphics[width=0.80\columnwidth]{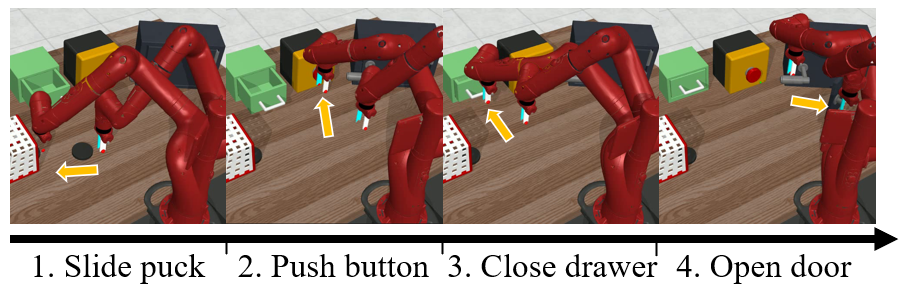}

    \caption{Multi-stage Meta-world task}
    \label{fig:multistage_metaworld}
        \vskip -0.1 in
\end{figure}
Given a demonstration $d^{\mathcal{T}}$ for task $\mathcal{T}
\sim \mathbb{T}_{\text{train}}$, we consider a model $\phi$ which maps $d^{\mathcal{T}}$ to a policy $\pi_{\mathcal{T}} = \phi(d^{\mathcal{T}})$. 
Then, the objective of one-shot imitation is to find the optimal model $\phi^*$ such that 
\begin{equation}
    \underset{\phi}{\text{argmax}} \left[\underset{\mathcal{T}' \sim \mathbb{T}_{\text{eval}}}{\mathbb{E}} \left[\text{E}(\mathcal{T}', \phi(d^{\mathcal{T}'}))\right]\right]
\end{equation}
where $\text{E}$ is an evaluation function for a policy $\phi(d^{\mathcal{T}'})$ against a new task $\mathcal{T}'$.

To deal with one-shot imitation upon various dynamics, we consider a problem setting of a Hidden Parameter Markov decision process without reward (HiP-MDP$\setminus\text{R}$) with a tuple $(\mathcal{S}, \mathcal{A}, \mathcal{P}, \gamma, \mathcal{H}, P_{\mathcal{H}})$. 
Here, $\mathcal{S}$ is a state space, $\mathcal{A}$ is an action space, $\mathcal{P}$ is a transition probability, $\gamma$ is a discount factor, and $\mathcal{H}$ is a hidden parameter space, where $h \in \mathcal{H}$ determines the dynamics  $s' \sim \mathcal{P}(s, a; h)$ for states $s, s' \in \mathcal{S}$ and action $a \in \mathcal{A}$. 
$P_{\mathcal{H}}$ is a prior distribution of $\mathcal{H}$.
Then, the objective of one-shot imitation and zero-shot adaptation across various dynamics is to find the optimal model $\phi^*$ such that 
\begin{equation}
\underset{\phi}{\text{argmax}}\left[
\underset{h \sim P_\mathcal{H}}{\mathbb{E}} \left[
\underset{\mathcal{T}' \sim \mathbb{T}_{\text{eval}}}{\mathbb{E}} \left[\text{E}_h({\mathcal{T}'}, \phi(d^{\mathcal{T}'}))\right]\right]\right]
\end{equation}
where $\text{E}_h$ is an evaluation function for a policy $\phi(d^{\mathcal{T}'})$ against task $\mathcal{T}'$ with fixed $h$ in a HiP-MDP$\setminus$R.

Regarding task definitions, we consider long-horizon, multi-stage tasks, in the form of $K$-staged tasks $\mathcal{T}_{0:K-1}$
(e.g., in Figure~\ref{fig:multistage_metaworld}, a 4-stage task with sequential subtasks, slide puck and others).
Note that $\mathcal{T}_{0:K-1}$ should be done in a correct order, e.g., slide puck, push button, close drawer, and then open door, specifying that the evaluation function $\text{E}_h$ yields some positive value for performing subtask $\mathcal{T}_{j}$ only when subtasks $\mathcal{T}_{0:j-1}$ have been completed.


\section{Our Approach}

\subsection{Overall Framework}
To enable the one-shot imitation and zero-shot adaptation in a non-stationary environment, we develop the OnIS framework, leveraging the semantic representation capability of a large scale vision-language pretrained model and compositionality of complex tasks. 

\begin{figure*}[t]
    \centering
    \includegraphics[width=0.90\textwidth]{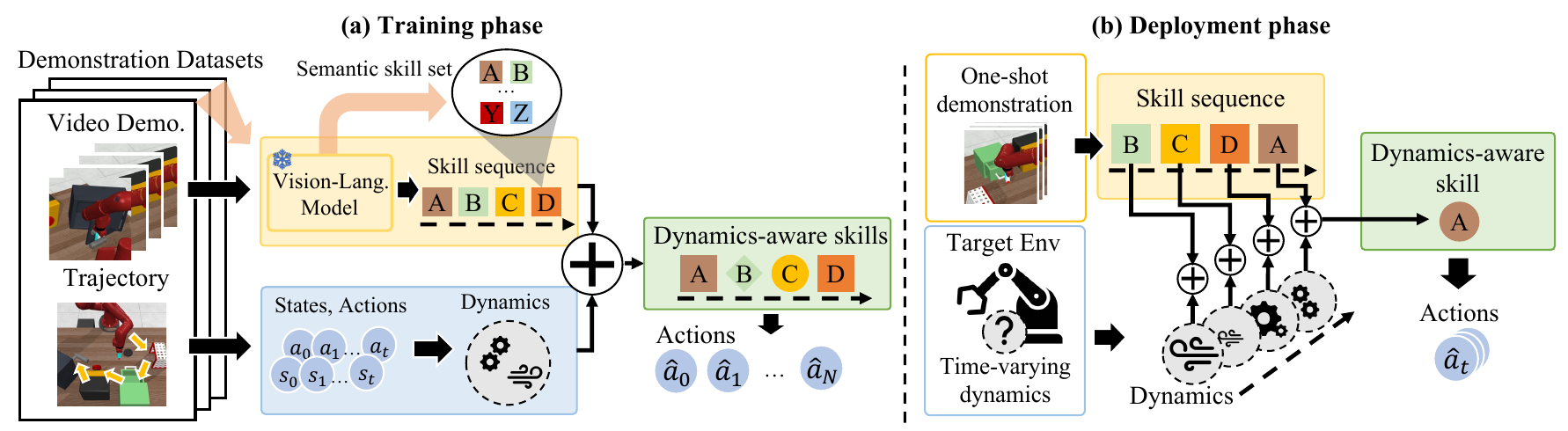}
    \caption{One-shot imitation and zero-shot adaptation in OnIS: 
    (a) In the training phase, a semantic skill set is established from video demonstrations in offline datasets by leveraging a pretrained vision-language model. For each demonstration in the dataset, semantic skill sequence and environment dynamics are first disentangled, and then they are combined to enable dynamics-aware skill transfer to different environments.
    (b) In the deployment phase, given a single demonstration for an unseen task, its semantic skill sequence can be immediately inferred (one-shot imitation), where each in the sequence adapts temporally to the hidden, time-varying dynamics in the environment (zero-shot adaptation). 
    }
    \label{fig:framework}
\end{figure*}

As illustrated in Figure~\ref{fig:framework}, the framework has two phases: (a) training phase and (b) deployment phase.  
In the training phase, we leverage task compositionality and pretrained vision-language models to extract semantic skills from video demonstrations. Specifically, we explore skill decomposition (task compositionality) for dealing with complex tasks where each task consists of sequential skills (achievable goals). Each task is decomposed and translated into a sequence of skills that are task-agnostically learned on offline datasets.  
From demonstrations for various tasks and diverse dynamics, the framework employs contrastive learning with video and text retrieval tasks to disentangle dynamics-invariant task-relevant features from the demonstrations, thus representing them in a vision-language semantic space, i.e., in the CLIP embedding space~\cite{radford2021learningCLIP}.  

At the same time, for dealing with adaptation across different dynamics, we develop a novel skill representation, dynamics-aware skills. The framework uses contrastive learning with dynamics reconstruction tasks to disentangle dynamics information from the trajectories in state and action pairs. 
These two individual contrastive learning procedures tend to decompose unstructured features contained in the demonstrations and trajectories into semantic skill sequences and environment dynamics, respectively.  
Then, to render a skill sequence adaptive to different dynamics, the decomposed skill sequence and dynamics are combined in training the skill transfer capability through a meta-learning procedure.  

(b) In the deployment phase, given a single video demonstration for an unseen task, the framework derives such a policy not only imitating the demonstrated expert behavior to complete the unseen task but also being capable of adapting to dynamics changes in the environment. In this phase, the framework uses only a single demonstration (one-shot imitation), and it enables rapid adaptation to dynamics changes (zero-shot adaptation) in a non-stationary environment, by making use of the knowledge constructed in the training phase.  

\begin{figure*}[t]
    \centering
    \includegraphics[width=0.89\textwidth]{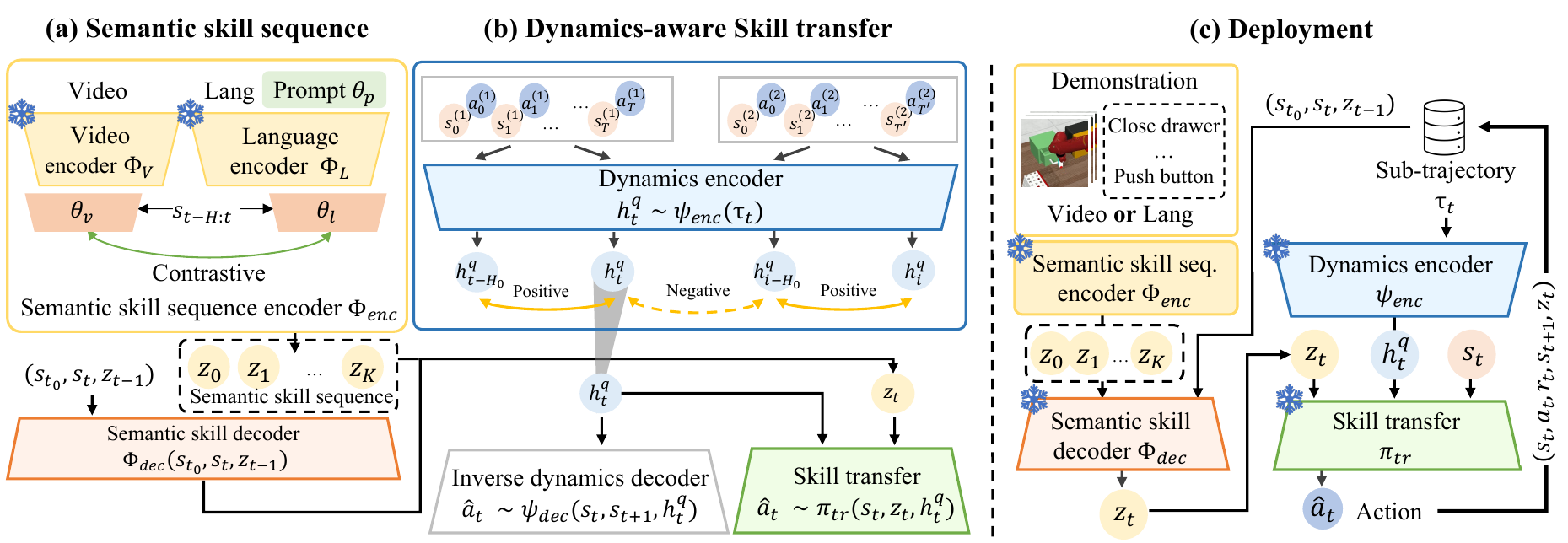}
    \caption{OnIS framework: In (a), the semantic skill sequence encoder $\Phi_{enc}$ and the semantic skill decoder $\Phi_{dec}$ are trained offline using the CLIP vision-language pretrained model, where $\Phi_{enc}$ translates video demonstrations to semantic skill sequences and is contrastively learned, and $\Phi_{dec}$ learns to infer an optimal skill (from a sequence) upon a state. 
    The language prompt $\theta_p$ is only used for the subtask-level instruction (S-OnIS) case, and the additional encoders $\theta_v, \theta_l$ are only used for episode-level instruction case (U-OnIS). 
    In (b), the skill transfer $\pi_{tr}$ and the dynamics encoder $\psi_{enc}$ are trained offline, where $\pi_{tr}$  learns to infer an action sequence optimized for the deployment setting from a given semantic skill sequence and inferred dynamics, and $\psi_{enc}$ learns to infer dynamics from sub-trajectories. These modules establish dynamics-aware skill transfer.  
    In (c), for a given demonstration, $\Phi_{enc}$ first infers a sequence of semantic skills, $\Phi_{dec}$ infers a current semantic skill, and $\psi_{enc}$ infers current dynamics in the non-stationary deployment environment. Then, $\pi_{tr}$ yields actions optimized through the current semantic skill and dynamics. 
    }
    \label{fig:methods}
\end{figure*}

For one-shot imitation, in the training phase, we exploit offline datasets of video demonstrations and their matched trajectories in state and action pairs, presuming the availability of such datasets from the environment.  
The contrastive learning procedures on the datasets and the meta-learning procedure together enable to construct the knowledge required for one-shot imitation in a non-stationary environment. 
The contrastive learning procedures are implemented in the framework, as a semantic skill sequence encoder and a dynamics encoder, and the meta-learning procedure is implemented as a skill transfer module. 
We explain these modules in Figure~\ref{fig:methods}, and Sections~\ref{section:semantic_skills_sequence} and~\ref{section:skill_transfer}.

\subsection{Learning Semantic Skill Sequences}\label{section:semantic_skills_sequence}

\textbf{Semantic skill sequence encoder.}
A semantic skill sequence encoder $\Phi_{enc}$ is responsible for mapping an expert demonstration $d$ to a sequence of semantic skills, in that $d$ is segmented into a sequence of dynamics-invariant behavior patterns (action sequences). Each pattern corresponds to a short sequence of expert actions and it can be described in a language instruction in the environment. 
Such an expert behavior pattern associated with some language instruction is referred to as a semantic skill, as for an expert behavior pattern, its associated language instruction is used to represent expert behaviors on the semantic embedding space of a vision-language aligned model. 

To implement the encoder $\Phi_{enc}$, we use the CLIP vision-language pretrained model and sample-efficient prompt learning techniques~\cite{zhou2022learningcoop}. 
Specifically, we assume that a dataset of expert trajectories $\mathcal{D} = \{\tau_1, \tau_2, \cdots, \tau_N\}$ contains language instruction data, in that each $T$-length trajectory $\tau = \{(s_1, v_1, l_1, a_1), \cdots, (s_T, v_T, l_T, a_T) \}$ consists of state $s$, visual observation $v$, language instruction $l$, and action $a$, where each language instruction is an element of the instruction set $\mathbf{L}$, similar to~\cite{zheng2022vlmbench}. 

Furthermore, we consider two distinct cases of language instructions annotated in the trajectories, one is the  
subtask-level instruction, (e.g.,``Push lever'' is annotated on the transitions for its relevant single subtask, 
as in \cite{pertsch2022crossdomainstar}), and the other is the episode-level instruction, (e.g., ``Push lever, open door and close the box'' is annotated on all the transitions in an episode, as in \cite{lisa, liu2022instructrl}). 
%

For the subtask-level instruction case, we assume that language instructions exist only for a small subset $\mathcal{D}_0$ of $\mathcal{D}$, 
and consider the CLIP model with a video encoder $\Phi_{V}$ and a language encoder $\Phi_{L}$ such that
\begin{equation}
    \begin{split}
        \Phi_{V}: v_{t: t+H} \mapsto z_v, \ \ \    
        \Phi_{L}: l_t \mapsto z_l.
    \end{split}
\end{equation}
These $\Phi_{V}$ and $\Phi_{L}$ encoders map a video demonstration $v_{t: t+H}$ and a language instruction $l_t$ into the same embedding space for timesteps $t$.
To construct $\Phi_{enc}$ based on $\Phi_{V}$ and $\Phi_{L}$ using a small number of samples, we use the language prompt~\cite{zhou2022learningcoop}.
Given  $v_{t:t+H}$ or $l_{t}$ as input $d$, the semantic skill sequence encoder $\Phi_{enc}$ learns to translate each input into a sequence of semantic skills by
\begin{equation}
    \begin{split}
	 \Phi_{enc}(d) =
	 \begin{cases}
        \underset{z \in z_\mathbf{L}}{\mathrm{argmax}}\{ \mathrm{sim} (z, \Phi_V(v_{t: t+H}))\}, &d=d_v \\
        \{\Phi_L(l_t ; \theta_p)\}, &d=d_l 
      \end{cases}
    \end{split}
    \label{eqn:semantic_skill_encoder_forward}
\end{equation}        
    where $d_v = (v_{t:t+H}: 0 \leq t \leq T')$,
         $d_l = (l_t: 0 \leq t \leq N)$,  
         $z_{\mathbf{L}} = \Phi_L(\mathbf{L}; \theta_p)$, and 
         $\mathbf{L}$ denotes a set of language instructions. 
Specifically, a language prompt $\theta_p$ in learnable parameters are trained via contrastive learning on positive pairs $(v_{t: t+H}, l_t)$ in $\mathcal{D}_0$,
where the contrastive loss is defined as $\mathcal{L}_{CON}(\theta_p) =$
\begin{equation}
    \begin{split}
    -\log \left(\frac{\mathrm{sim}(\Phi_V(v_{t:t+H}) , \Phi_L(l_t ; \theta_p))}{\sum_{l_t \neq l \in \mathbf{L}}\mathrm{sim}(\Phi_V(v_{t: t+H}), \Phi_L(l; \theta_p))}\right).
    \end{split}
    \label{loss:prompt}
\end{equation}
The similarity of latent vectors $z$ and $z'$ is calculated by
\begin{equation}
    \mathrm{sim}(z, z') = \frac{1}{\alpha} \exp\left(\frac{\langle z, z' \rangle}{\| z \| \| z' \|}\right)
    \label{loss:sup}
\end{equation}
where $\alpha$ is a temperature coefficient.
As such, for the subtask-level instruction case, the semantic skill sequence encoder $\Phi_{enc}$ can be established from the pretrained visual and language encoders $\Phi_{V}$, $\Phi_{L}$ through prompt-based contrastive learning on a small set of video and text retrieval samples. 

In the case of episode-level instruction, all the transitions in an episode in $\mathcal{D}$ are associated only with a single instruction without detailed subtask-level instructions. Thus, we adapt unsupervised skill learning techniques in~\cite{lisa} with contrastive learning on video features.  
Accordingly, $\Phi_{enc}(d)$ learns to translate a video demonstration $d=v_{0: T'}$ or language instruction $d=l$ to a sequence of semantic skills by
\begin{equation}
\begin{split}	
	\begin{cases}
        \{\underset{z \in z_{\mathbf{L}}}{\mathrm{argmax}} \{ \mathrm{sim} (z, \theta_v (\Phi_V(v_{0: T'}), s_{t-H: t}) \} \}, &d = v_{0: T'} \\
		\{\theta_l ( \Phi_L(l), s_{t-H: t}) \}, &d = l
	\end{cases}
 \end{split}
    \label{eqn:uonis_semantic_skill_encoder_forward}
\end{equation}
where $z_{\textbf{L}} = \theta_l (\Phi_L(\mathbf{L}), s_{t-H:t})$.
Similar to the skill predictor used in~\cite{lisa}, two additional encoders $\theta_v$ and $\theta_l$ such as
\begin{equation}
\begin{split}
&\theta_v: (\Phi_V(v_{0: T'}), s_{t-H: t}) \mapsto z_v \\
&\theta_l: (\Phi_L(l), s_{t-H: t}) \mapsto z_l
\end{split}
\label{eqn:uonis_layer}
\end{equation}
are trained.
Specifically, the encoder $\theta_v$ is contrastively learned on positive pairs $(v_{0: T'}, l)$ in $\mathcal{D}$, where the contrastive loss in~\eqref{loss:prompt} is rewritten as 
$\mathcal{L}_{CON}(\theta_v)$ =
\begin{equation}
    \begin{split} -\log \left(\frac{\mathrm{sim}(\theta_v(\Phi_V(v_{0: T'}), s_{t-H: t}), z_l)}{\sum_{z_l \neq z \in z_{\mathbf{L}}}\mathrm{sim}(\theta_v(\Phi_V(v_{0: T'}), s_{t-H: t}), z)}\right)
    \end{split}
    \label{loss:unsup}
\end{equation}
for $z_l = \theta_l(\Phi_L(l), s_{t-H: t})$ and $z_{\textbf{L}} = \theta_l (\Phi_L(\mathbf{L}), s_{t-H:t})$.
The encoder $\theta_l$ is trained via behavior cloning that maximizes the mutual information between the actions and semantic skills, where the loss is defined as
\begin{equation}
    \mathcal{L}_{BC}(\theta_l, f) = \mathbb{E} [\| a_t - f(s_t, z_l) \|]
\end{equation}
\label{eqn:unsup_bc}
for the action reconstruction model $f: (s_t, z_l) \mapsto a_t$.

\textbf{Semantic skill decoder.}
Given a sequence of semantic skills, the semantic skill decoder $\Phi_{dec}$ is responsible for obtaining a semantic skill for a current state. 
$\Phi_{dec}$ is implemented as a binary model to determine whether the current semantic skill is terminated or not, similar to the task specification interpreter in~\cite{xu2018neuralntp}, i.e., 
\begin{equation}
    \Phi_{dec}: (s_{t_0}, s_{t}, z_t) \mapsto \{0, 1\}   
\end{equation}
for a semantic skill $z_t$ being currently executed, the initial state $s_{t_0}$ for $z_t$,  and a current state $s_t$. 
$\Phi_{dec}$ is learned on $\mathcal{D}$ with binary cross entropy (BCE),
\begin{equation}
    \label{termination:bce}
    \mathcal{L}_{BCE}(\Phi_{dec}) = \text{BCE}(\mathbbm{1}_{z_t\neq z_{t+1}}, \Phi_{dec}(s_{t_0}, s_t, z_t))
\end{equation}
where $z_t = \Phi_{v}(v_{t:t+H})$. 

As illustrated in Figure~\ref{fig:methods}(a), the semantic skill sequence encoder $\Phi_{enc}$ is trained through contrastive learning to extract the inherent semantic skills in the video. The semantic skill decoder $\Phi_{dec}$, which is basically a binary classifier, is conditioned on a skill sequence and is trained to predict the appropriate semantic skill at the current timestep.

\subsection{Learning Dynamics-aware Skill Transfer}
\label{section:skill_transfer}
Given a semantic skill sequence encoder $\Phi_{enc}$, the skill transfer module $\pi_{tr}$ is responsible for transferring a semantic skill $z_t = \Phi_{enc}(v_{t: t+H})$ to an action sequence adapted for environment dynamics. That is, 
\begin{equation}
    \pi_{tr}: (s_t, z_t, h^q_t) \mapsto a_t
    \label{eqn:skill_transfer}
\end{equation}
for a current state $s_t$, a semantic skill $z_t$, and the dynamics embedding $h^q_t=\psi_{enc}(\tau_t)$.
For $H_0$-length sub-trajectories $\tau_t = (s_{t-H_0}, a_{t-H_0}, ..., s_{t-1}, a_{t-1})$, 
the dynamics encoder $\psi_{enc}$ takes it as input, and maps to a quantized vector $h^q_t$, i.e., 
\begin{equation}
    \psi_{enc} = \mathbf{q} \circ \psi^c_{enc}: \tau_t \mapsto h^q_t
    \label{eqn:quantization}
\end{equation}
where $\psi^c_{enc}$ maps $\tau_t$ to a continuous latent vector.
We use the vector quantization operator $\mathbf{q}$ to avoid the posterior collapse~\cite{NIPS2017vqvae}.
The output of $\mathbf{q}$ is the nearest vector among the learnable parameters, called codebook, $h^q \in \{h^{q_1}, \cdots, h^{q_M} \}$. Then, the quantized dynamics embedding $h^q_t$ is obtained by
\begin{equation}
    h^q_t = \mathbf{q}(\psi^c_{enc}(\tau_t)) = \underset{j \in \{1, \cdots, M\}}{\textnormal{argmin}} \{ \| \psi^c_{enc}(\tau_t) - h^{q_j}\| \}.
    \label{eqn:vector_quantization}
\end{equation}
For a semantic skill $z_t = \Phi_{enc}(v_{t:t+H})$ and dynamics embedding $h^q_t = \psi_{enc}(\tau_t)$,
the skill transfer module and dynamics encoder are jointly trained by minimizing the behavior cloning (BC) loss $\mathcal{L}_{BC}(\pi_{tr}, \psi_{enc}) = $
\begin{equation}
    \begin{split}
     \mathbb{E}_{s_T, a_T \sim \mathcal{D}}\left[\|a_t - \pi_{tr}(s_t, z_t, \psi_{enc}(\tau_t))\|^2\right].
    \end{split}
    \label{equ:L_BC}
\end{equation}
Furthermore, to disentangle task-irrelevant dynamics from sub-trajectories,  we also use contrastive learning on sub-trajectories in various dynamics.
%
%
Specifically, suppose that a batch of $H_0$-length sub-trajectories $\{\tau_{t_i}\}_{1 \leq i \leq N}$ 
contains one positive sample $\{(\tau_{t_j}, \tau_{t_k}) \}$ which comes from the same trajectory starting at different timesteps. 
Then, $\psi_{enc}$ is learned on the positive and negative pair, where the contrastive loss is defined as 
$\mathcal{L}_{CON}(\psi_{enc}) = $ 
\begin{equation}
\begin{split}
    &-\log \left(\frac{\mathrm{sim}(\psi_{enc}(\tau_{t_j}), \psi_{enc}(\tau_{t_k}))}{\sum_{i \neq i'} \mathrm{sim}(\psi_{enc}(\tau_{t_i}), \psi_{enc}(\tau_{t_{i'}}))}\right).
\end{split}
    \label{equ:L_CON}
\end{equation}
To maximize the mutual information between the inferred embedding $h^q_t$ and underlying dynamics, we adopt reconstruction-based feature extraction by using the inverse dynamics decoder $\psi_{dec}: (s_t, s_{t+1}, h_t^q) \mapsto a_t$, where the action reconstruction loss is defined as
\begin{equation}
    \begin{split}
    &\mathcal{L}_{REC}(\psi_{enc}, \psi_{dec}) \\
    &= \mathbb{E}_{t-H_0 \leq i < t}\left[\|a_i - \psi_{dec}(s_i, s_{i+1}, \psi_{enc}(\tau_t)\|^2\right].
    \end{split}
    \label{equ:L_REC}
\end{equation}

\begin{algorithm}[t]
    \caption{Learning to transfer skills}
    \begin{algorithmic}[1]
    \STATE Semantic skill sequence encoder $\Phi_{enc}$
    \STATE Dynamics encoder $\psi_{enc}$, Inverse dynamics decoder $\psi_{dec}$
    \STATE Skill transfer module $\pi_{tr}$, Dataset $\mathcal{D}$
    \REPEAT
    \STATE Sample a batch $\{(\tau_{t_i}, v_{t_i:t_i+H})\}_i \sim \mathcal{D}$
    \STATE $\{z_{t_i}\}_i = \Phi_{enc}(\{v_{t_i:t_i+H}\}_i)$
    \STATE \textit{/* Calculate loss with} $\{(\tau_{t_i}, z_{t_i})\}$ \textit{*/} 
    \STATE $\text{loss}_{bc} \gets \mathcal{L}_{BC}(\pi_{tr}, \psi_{enc})$ using~\eqref{equ:L_BC}
    \STATE $\text{loss}_{con} \gets \mathcal{L}_{CON}(\psi_{enc})$ using~\eqref{equ:L_CON}
    \STATE $\text{loss}_{rec} \gets \mathcal{L}_{REC}(\psi_{enc}, \psi_{dec})$ using~\eqref{equ:L_REC}
    \STATE $\psi_{enc} \gets \psi_{enc} - \nabla_{\psi_{enc}}(\text{loss}_{bc}+\text{loss}_{con}+\text{loss}_{rec})$
    \STATE $\psi_{dec} \gets \psi_{dec} - \nabla_{\psi_{dec}} \text{loss}_{rec}$
    \STATE $\pi_{tr} \gets \pi_{tr} - \nabla_{\pi_{tr}} \text{loss}_{bc}$
    \UNTIL{converge}
	\end{algorithmic}
	\label{alg:skill_transfer}
 
\end{algorithm}
\vskip -0.1in
Overall, the skill transfer module $\pi_{tr}$, the dynamics encoder $\psi_{enc}$, and the inverse dynamics decoder $\psi_{dec}$ are jointly trained to minimize the losses $\mathcal{L}_{BC}, \mathcal{L}_{CON}$ and $\mathcal{L}_{REC}$, where these modules are presented in Figure 3(b).
During the deployment phase, the skill transfer module acts as a policy network that determines actions ($\hat{a}_t$) upon 
 the skill and dynamics embeddings ($z_t$ and $h^q_t$) along with the current state ($s_t$), as illustrated in Figure 3(c).

Algorithm~\ref{alg:skill_transfer} lists the learning procedures for skill transfer in Section~\ref{section:skill_transfer}.

\section{Evaluations}\label{sec:evaluations}
In this section, we evaluate the performance of our OnIS framework under various configurations of non-stationary environments. 

\noindent \textbf{Environment settings.}
For evaluation, we devise multi-stage robotic manipulation tasks, namely multi-stage Meta-world, using the Meta-world simulated benchmark~\cite{yu2020metaworld}, where each multi-stage task is composed of a sequence of existing Meta-world tasks (subtasks). An agent manipulates a robot arm using actions such as slide puck, close drawer, etc., to achieve certain objectives.
For emulating non-stationary environments, we exploit kinematic parameters affecting the dynamics of robotic manipulation, which are not explicitly revealed through states or videos. 
This non-stationarity is consistent with the prior works in robotics~\cite{kumar_rma, kumar_motor}.

\noindent \textbf{Offline datasets.} 
To generate the dataset that covers diverse dynamics, we implement rule-based expert policies to collect trajectories from 13 different stationary environment conditions for each subtask, where each condition corresponds to a specific kinematic configuration.
We collect $3120$ episodes (240 for each environment condition), and we annotate $24$ episodes (0.76\% of our dataset) with subtask-level language instructions. This annotated data is only used to train the semantic skill sequence encoder, especially for S-OnIS (in Section~\ref{subsec:impl}).  

\subsection{OnIS Implementation}\label{subsec:impl}
We implement our OnIS framework using the opensource project Jax~\cite{jax2018github}. 
OnIS is composed of 4 modules: semantic skill sequence encoder $\Phi_{enc}$, skill decoder $\Phi_{dec}$, dynamics encoder $\psi_{enc}$, and skill transfer module $\pi_{tr}$.
We use the Transformer-based CLIP pretrained model~\cite{radford2021learningCLIP} to implement the sequence encoder. 
Specifically, we implement two versions of OnIS: one with supervised semantic skills (S-OnIS) and the other with unsupervised semantic skills (U-OnIS). The former S-OnIS corresponds to subtask-level instructions where contrastive learning for semantic skills is driven by the supervision of subtask-level instructions (with sub-trajectory and skill pairs). 
The latter U-OnIS corresponds to episode-level instructions where 
contrastive learning for semantic skills is driven by the supervision of episode-level instructions (without sub-trajectory and skill pairs).

\subsection{Baselines}
\begin{itemize}[leftmargin=*, noitemsep]
    \item BC-Z \cite{jang2022bcz} is a state-of-the-art imitation learning framework, which utilizes both video demonstrations and language instructions.
    To extract semantic skill information from multi-modal data, BC-Z is trained to maximize the similarity between a visual and language embedding encoded by the pretrained language model.
    \item Decision Transformer (DT)~\cite{decision_transformer}  is a Transformer-based model tailored for imitation learning. 
    Considering that DT is not originally meant for multi-modal data, we modify DT to have state vectors in different modalities, along with either a video demonstration or language instruction. 
    We use DT to compare OnIS with conventional imitation learning approaches which do not consider skill semantics.
    \item SPiRL \cite{spirl} is a state-of-the-art unsupervised skill-based RL algorithm, which embeds sub-trajectories into the latent skill space. 
    Similar to DT, we modify SPiRL to handle either a video demonstration or language instruction in a single framework.
\end{itemize}


\subsection{One-shot Imitation Performance}\label{subsec:oneshot_imitation_performance}
Tables~\ref{main:visual} and~\ref{main:language} compare the one-shot imitation performance in task success rates achieved by our framework (U-OnIS, S-OnIS) and other baselines (DT, SPiRL, BC-Z), given two distinct scenarios in terms of modality with a single input in either a video demonstration or language instruction. 
As shown in Figure~\ref{fig:methods}(c), all the learned modules in the framework freeze when being deployed in a target environment, and they are immediately evaluated against various demonstrations and time-varying dynamics conditions.
%
%

Specifically, we measure the one-shot imitation performance in task success rates for multi-stage tasks with $K$ sequential objectives ($K=1,2,4$) upon several dynamics change levels (stationary, low, medium, and high).
Regarding baselines, DT and SPiRL variants for leveraging video demonstrations are denoted as V-DT and V-SPiRL, and those for leveraging language instructions are denoted as L-DT and L-SPiRL.
Due to its low performance, V-DT is not included in the comparison.
\begin{table}[h]
    \caption
    {One-shot imitation performance for video demonstration: for multi-stage Meta-world tasks, the performance in task success rates by our U-OnIS, S-OnIS and other baselines is measured against various test conditions on $K$ sequential objectives in a task ($K=1,2,4$) and dynamics change levels (stationary, low, medium, and high). 
    }
    \vskip 0.1in
    \footnotesize
    \centering
    \resizebox{0.9\columnwidth}{!}
    {
        \begin{tabular}{c|c||c|c||c|c}
        \hline
        \multicolumn{1}{c|}{$K$}  & \multicolumn{1}{c||}{Non-st.} 
        & \multicolumn{1}{c|}{V-SPiRL} & \multicolumn{1}{c||}{BC-Z} & \multicolumn{1}{c|}{U-OnIS} & \multicolumn{1}{c}{S-OnIS} \\
        \hline

        \multirow{4}{*}{1} &
        Stationary      & 61.48\% & 75.00\% & 95.80\% & \textbf{100.0\%} \\ \cline{2-6}
        & Low           & 48.23\% & 48.33\% & 86.28\% & \textbf{94.55\%} \\  
        & Medium        & 45.22\% & 44.29\% & 84.09\% & \textbf{94.34\%} \\  
        & High          & 39.09\% & 37.63\% & 81.10\% & \textbf{90.49\%} \\  
        \hline

        \multirow{4}{*}{2} & 
        Stationary      & 61.75\% & 71.08\% & 67.50\% & \textbf{100.0\%} \\ \cline{2-6}
        & Low           & 41.66\% & 42.85\% & 75.75\% & \textbf{85.66\%} \\
        & Medium        & 43.39\% & 47.61\% & 73.07\% & \textbf{84.29\%} \\  
        & High          & 30.33\% & 22.09\% & 66.00\% & \textbf{81.62\%} \\ 
        \hline

        \multirow{4}{*}{4} & 
        Stationary      & 47.44\% & 21.01\% & 64.38\% & \textbf{91.67\%} \\ \cline{2-6}
        & Low           & 27.31\% & 14.38\% & 50.03\% & \textbf{74.95\%} \\
        & Medium        & 20.82\% & 12.79\% & 49.22\% & \textbf{74.89\%} \\
        & High          & 15.54\% & 11.13\% & 49.82\% & \textbf{70.19\%} \\  
        \hline
        \end{tabular}
    }
    \label{main:visual}
    \vskip -0.1in
\end{table}

\begin{table}[h]
    \caption
    {
    One-shot imitation performance for language instruction 
    }
    \vskip 0.1in
    \footnotesize
    \centering
    \resizebox{\columnwidth}{!}
    {
        \begin{tabular}{c|c||c|c|c||c|c}
        \hline

        \multicolumn{1}{c|}{$K$}  & \multicolumn{1}{c||}{Non-st.} 
        & \multicolumn{1}{c|}{L-DT} & \multicolumn{1}{c|}{L-SPiRL} & \multicolumn{1}{c||}{BC-Z} & \multicolumn{1}{c|}{U-OnIS} & \multicolumn{1}{c}{S-OnIS} \\
        \hline

        \multirow{4}{*}{1} &
        Stationary            & 92.66\% & 100.0\% & 100.0\% & 87.70\% & \textbf{100.0\%} \\ \cline{2-7}
        & Low           & 62.49\% & 54.29\% & \textbf{96.61\%} & 90.40\% & 95.00\% \\  
        & Medium        & 53.84\% & 50.71\% & 75.32\% & 81.29\% & \textbf{89.04\%} \\  
        & High          & 20.19\% & 47.75\% & 68.91\% & 82.19\% & \textbf{88.50\%} \\  
        \hline

        \multirow{4}{*}{2} & 
        Stationary            & 58.33\% & 60.08\% & 76.25\% & 74.50\% & \textbf{95.00\%} \\ \cline{2-7}
        & Low           & 47.12\% & 29.35\% & 45.27\% & 60.44\% & \textbf{83.17\%} \\
        & Medium        & 33.17\% & 34.34\% & 35.00\% & 69.24\% & \textbf{78.50\%} \\  
        & High          & 30.28\% & 20.51\% & 25.52\% & 66.67\% & \textbf{77.27\%} \\ 
        \hline

        \multirow{4}{*}{4} & 
        Stationary            & 37.73\% & 43.16\% & 40.30\% & 67.71\% & \textbf{87.50\%} \\ \cline{2-7}
        & Low           & 22.11\% & 22.83\% & 20.23\% & 59.64\% & \textbf{76.60\%} \\
        & Medium        & 20.43\% & 15.50\% & 20.60\% & 56.68\% & \textbf{71.39\%} \\
        & High          & 8.90\% & 16.39\% & 11.13\% & 54.82\% & \textbf{66.92\%} \\  
        \hline
        \end{tabular}
    }
    \label{main:language}
        \vskip -0.1in
\end{table}

\textbf{Overall performance.} 
As shown in Table~\ref{main:visual}, our U-OnIS and S-OnIS yield higher performance consistently than the baselines for all video demonstration cases. Specifically, S-OnIS outperforms the most competitive baseline SPiRL by a significant margin $38.25\%\sim54.64\%$ on average.
The performance gap between S-OnIS and the baselines increases for larger $K=4$ where a task consists of 4 subtasks, compared to $K=1,2$. This specifies that our approach is competitive in learning on a relatively long-horizon demonstration that performs multi-stage tasks.

\textbf{Consistency in multi-modality.}
Similar to the video demonstration scenario, in Table~\ref{main:language}, U-OnIS and S-OnIS show competitiveness for the language instruction scenario.
When Tables~\ref{main:visual} and~\ref{main:language} are compared, we observe that
U-OnIS and S-OnIS maintain high performance consistently for the two distinct scenarios, showing $5.69\%$ and $3.17\%$ performance differences (between the language and video scenarios) on average between them.
Unlike OnIS, although BC-Z explores multi-modality, it achieves inconsistent performance between the two scenarios, showing a $16.01\%$ performance drop for the video demonstration scenario from the language instruction scenario. 
This inconsistent performance pattern between different modalities can be also found in~\cite{jang2022bcz}.
While our approach exploits task compositionality with shareable subtasks and semantic skills, BC-Z tends to transform a whole video input into an individual task; thus, BC-Z can degrade significantly for new video demonstrations. 
%

\textbf{Generalization to non-stationarity.}
As shown in Tables~\ref{main:visual} and~\ref{main:language}, our U-OnIS and S-OnIS show more robust performance than the baselines for all non-stationary settings (including low, medium, and high).
Specifically, in Table~\ref{main:language}, S-OnIS achieves a relatively small drop of $9.49 \% \sim 18.13 \%$ on average in non-stationary environments compared to the stationary.
The most competitive baseline, BC-Z shows a large drop of $19.72\% \sim 57.02\%$ on average in the same condition.
In OnIS, the separate embedding spaces for semantic skills and dynamics are used to address time-varying dynamics. Without that decoupled embedding strategy, BC-Z can experience low-performance skills, which are entangled with certain dynamics in training environments.


\textbf{U-OnIS and S-OnIS.}
As presented above, while both U-OnIS and S-OnIS outperform the baselines, S-OnIS shows better performance than U-OnIS with an average margin of $16.63\%$. 
In our framework with efficient prompt learning, the ability of semantic skill learning can be improved by a small annotated dataset (i.e., 0.76\% of total trajectories of 3120 episodes) with subtask-level instructions.   

\begin{figure}[h]
    \centering
        \includegraphics[width=0.38\textwidth]{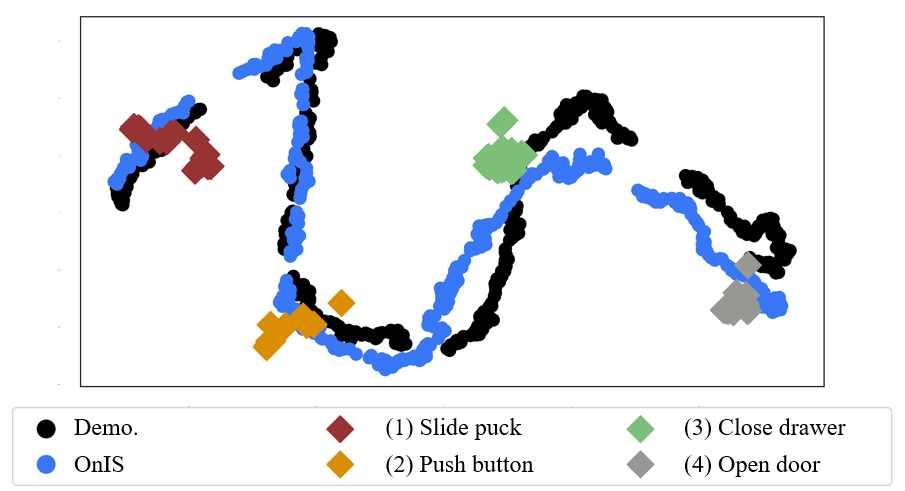}
    \caption{Semantic embedding correspondence for a demonstration and one-shot imitation run}
    \label{fig:an1}
\end{figure}

\textbf{Semantic embedding correspondence.} 
\label{expl:semantic_embedding_correspondence}
Given a video demonstration, in Figure \ref{fig:an1}, we visualize its embeddings (as black-colored circles, Demo.) and the embeddings by its respective episodic run by U-OnIS (as blue-colored circles, OnIS) in a multi-modal semantic space. As observed, two streams of the embeddings construct a similar path, indicating their temporal correspondence in semantic skills in an episode. 
We also visualize semantic skill embeddings (as rhombuses of different colors) of sub-trajectories in our dataset, which are matched with the 4 subtasks in the demonstration. The path (OnIS) turns out to follow these semantic skill embeddings. This correspondence specifies that a policy achieved by one-shot imitation can perform skills, which are required to complete the subtasks in the demonstration, in the correct sequence. 
%


\subsection{Ablation Study}

\textbf{Temporal contrast.}
Table~\ref{table:temporal_contrastive} evaluates our positive pair sampling strategy in the dynamics encoder, 
where Fixed$\pm N$ and Random$\pm N$ denote sampling on $N$-fixed interval and 
and random sampling within $N$ interval, respectively, in a trajectory.  
%
%
Specifically, our framework employs Random$\pm$T, where T is the trajectory length.  
With Random$\pm$T, the framework shows $3.75\% \sim 18.25\%$ higher performance than others. This is because pairs sampled in a whole episode contain a wider range of state variations with respect to dynamics. 
\begin{table}[h]
    \centering
    \caption{Effect by temporal contrast}
    \vskip 0.1in
    \resizebox{\columnwidth}{!}
    {
        \begin{tabular}{c||c|c|c|c}
    	\hline
    	Dynamics & Fixed$\pm$1 & Fixed$\pm$10 & Random$\pm$10 & Random$\pm$T \\
        \hline
        Unseen Stationary & 49.50\% & 54.25\% & 55.00 \% & 63.25\% \\ 
        Non-Stationary & 39.75\% & 44.50\% & 63.25\% & 62.50\% \\ 
        \hline
    	\end{tabular}
    }
    
\label{table:temporal_contrastive}

    \vskip -0.1in
\end{table}

\begin{figure}[h]
    \centering
    \includegraphics[width=0.30\textwidth]{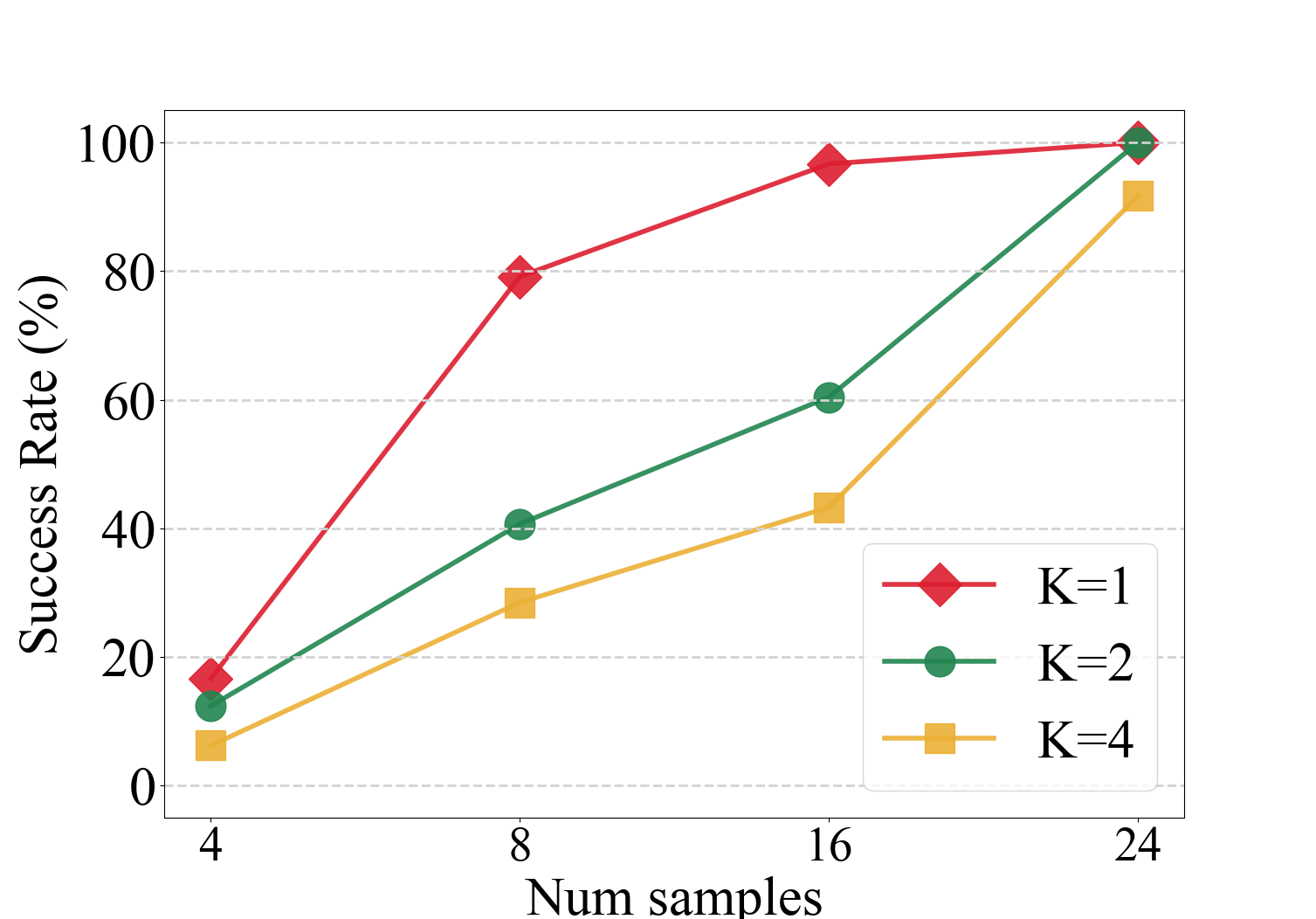}
    \caption{Effect by annotated sample size: the x-axis denotes the number of annotated samples used for training the semantic skill sequence encoder in S-OnIS, and the y-axis denotes the one-shot imitation performance by S-OnIS.}
    \label{fig:prompt_sample_efficiency}
    \vskip - 0.1 in
\end{figure}

\textbf{Annotation sample size.}
Figure~\ref{fig:prompt_sample_efficiency} evaluates S-OnIS with respect to different sample sizes (4, 8, 16, 24) used for training the prompt (in~\eqref{loss:prompt}).
As expected, the extremely low data regime of 4 samples rarely solves even a single task ($K=1$), but the default S-OnIS implementation with 24 samples achieves robust performance for all $K=1,2,4$. 
%
This shows the efficiency of prompt-based contrastive learning used for the semantic skill sequence encoder.
%
%
%
\subsection{Use Cases}
\textbf{Noisy video demonstration.}
To evaluate the robustness of our framework on noisy video demonstrations, we construct different scenarios: 
(1) Color jitter, 
(2) Gaussian noise, 
(3) Cutout. 
Table~\ref{noisyvisual} shows the performance of 4-stage tasks for the scenarios. 
The baselines are highly vulnerable to noisy video demonstrations, but OnIS is robust to them. 
In the training phase, OnIS employs video and text retrieval tasks with a vision-language model, tending to extract semantic information from visual observations. This renders OnIS robust to noisy demonstrations, especially when noise does not alter underlying semantics.  
\begin{table}[h]
	\centering
 	\caption{Noisy video demonstrations}
    \vskip 0.1in
    \resizebox{0.9\columnwidth}{!}
    {
        \begin{tabular}{c||c|c||c|c}
    	\hline
    	Noise type & V-SPiRL & BC-Z & U-OnIS & S-OnIS \\
        \hline
        w/o noise & 47.44\% & 21.01\% & 64.38\% & 91.67\% \\
        \hline
        Color jitter    & 0.00\% & 2.78\% & 35.19\% & 68.05\%\\ 
        Gaussian noise  & 0.00\% & 0.00\% & 25.42\% & 20.83\% \\
        Cutout          & 0.00\% & 0.00\% & 27.78\% & 8.33\% \\ 
        \hline
    	\end{tabular}
    }
    \label{noisyvisual}
    
    \vskip -0.1in
\end{table}

\begin{figure}[h]
    \centering
    \includegraphics[width=0.48\textwidth]{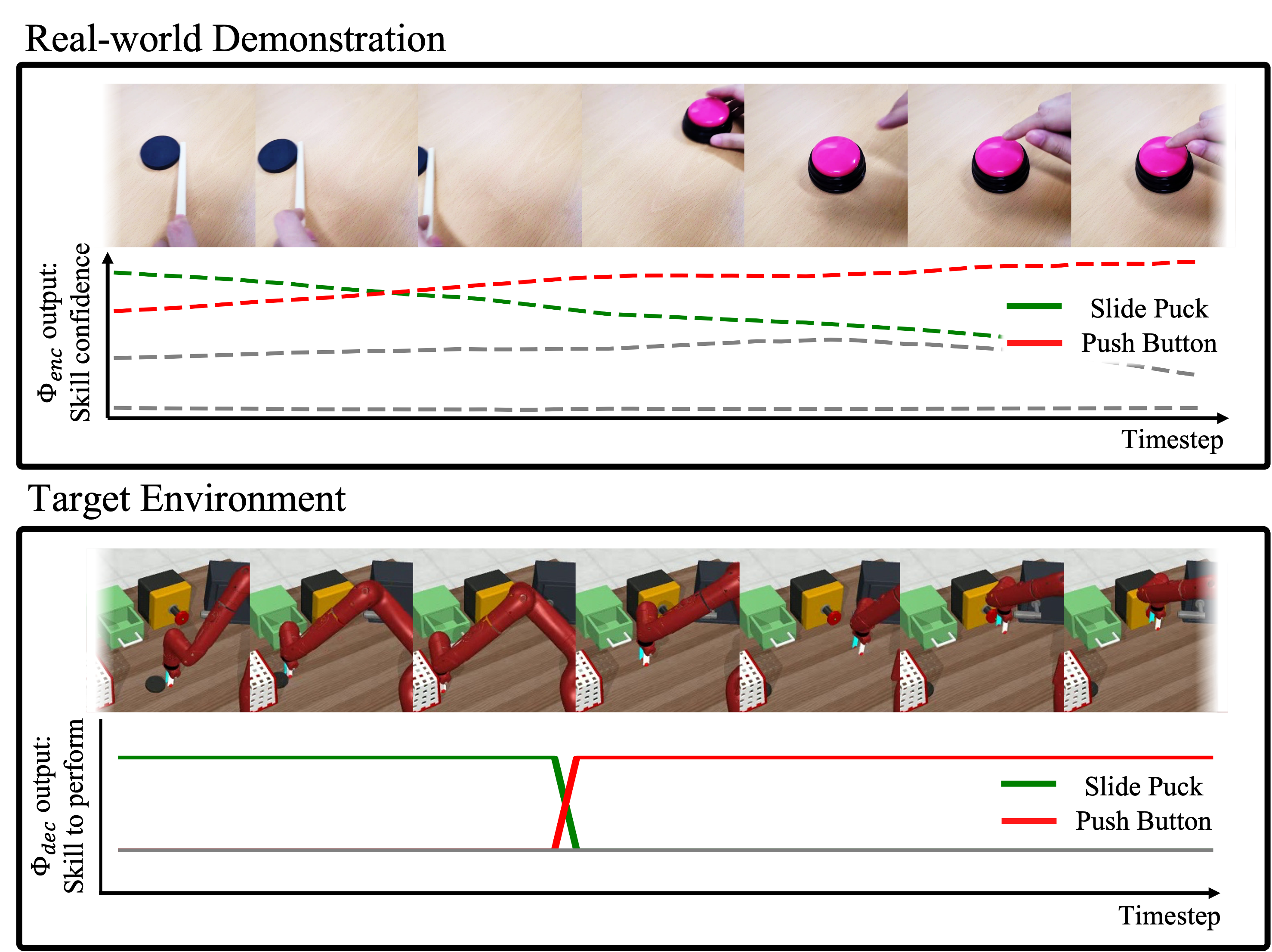}
    \vskip -0.1 in
    \caption{Real-world demonstrations}
    \label{fig:realworld}
    \vskip -0.1 in
\end{figure}

\textbf{Real-world video demonstration.} 
Here, we test our framework with real-world video demonstrations, where we use a few real-world video clips, e.g., slide puck and push button at the top of Figure~\ref{fig:realworld}.
%
%
The graph associated with the real-world video clips specifies the inference for pretrained semantics skills over timesteps, yielded by our semantic skill sequence encoder. The inference results state that the encoder is able to extract correct semantic skills from the input clips to perform respective robotic manipulation tasks, exploring the domain-invariant semantic knowledge tuned on the pretrained CLIP visual encoder.
At the bottom of the figure, we present the semantic skill execution over timesteps in the target environment, where each semantic skill for a certain timestep is determined by the semantic skill decoder and is then translated into an optimized action sequence through our skill transfer module. 

\textbf{Instruction variation.} 
We evaluate our framework with language instruction variants, which are not used for training the semantic skill sequence encoder. For this, we use several instruction variations with 
 verb replacement, modifier extension, and verbose instructions.
As shown in Table~\ref{table:novellanguage}, both U-OnIS and S-OnIS achieve robust performance against unseen language instructions with variations. This is because the sequence encoder does not directly map an instruction in the embedding space through the CLIP language encoder, but translates it to its respective semantic skill by using its similarity to learned semantic skills. 
\begin{table}[h]
    \centering
    \caption{Language instruction variations}
    \vskip 0.1in
    \resizebox{0.68\columnwidth}{!}
    {
    \begin{tabular}{c||c|c}
    \hline
        Variations & U-OnIS & S-OnIS \\
        \hline
        Seen instruction & 70.01\% & 93.50\% \\
        \hline
        Verb replacement & 70.01\% & 93.50\% \\ 
        Modifier extension & 70.01\% & 93.50\%  \\
        Verbose instruction & 40.74\% & 54.25\% \\ 
        \hline
    \end{tabular}
    }
	\label{table:novellanguage}
 
    \vskip -0.1in
\end{table}

\section{Related Works}
\textbf{One-shot imitation learning.} 
To adapt to a new task with a single demonstration, one-shot imitation learning has been investigated by extracting task information from state action sequences~\cite{NIPS2017_ba386660oneshotimitation, yu2018oneshothierarchicalrl} or 
from video demonstrations~\cite{Yu-RSS-18-daml, Mosaic, pmlr-v155-xu21dvisualimitationeasy}.
For example, \cite{Yu-RSS-18-daml} exploited human video demonstrations to learn robotic manipulation tasks, focusing on the generalization ability to unseen tasks in video demonstrations, and \cite{Mosaic} adopted contrastive learning schemes for cross-domain one-shot imitation. 
We share the same direction to one-shot imitation with these prior works, but we address specific one-shot imitation issues for a non-stationary environment, aiming to achieve one-shot imitation for unseen long-horizon tasks and zero-shot adaptation across various dynamics.  

\textbf{Skill-based RL.} 
To tackle long-horizon multi-stage tasks, skill-based RL techniques have been explored along with hierarchical learning frameworks. 
\cite{eysenbach2018diversity} explored unsupervised skill learning in a hierarchical framework without requiring explicit reward signals from the environment.
Recently, in~\cite{spirl,simpl}, with the availability of diverse expert trajectory datasets in multi-task environments, skill embedding techniques in offline settings were further investigated to accelerate online policy learning for long-horizon tasks. 
We also use skill embedding techniques to tackle multi-stage long-horizon tasks, but adapt them with a vision-language pretrained model to exploit semantics in skills and demonstrations.  Our OnIS is the first framework using semantic skill embeddings in the CLIP space to enable one-shot imitation and zero-shot adaptation. 

\textbf{Multi-modality for imitation learning.} 
Recently, several imitation learning approaches using large scale pretrained models have been introduced.
\cite{hiveformer} used a pretrained language model to tackle 
instruction-following tasks in a vision-based environment, and \cite{liu2022instructrl, shridhar2022cliport} used a pretrained multi-modal model to encode both instruction and visual observations for better scaling and generalization ability.
\cite{jang2022bcz} presented a vision-language joint learning method for one-shot imitation learning which can utilize both video demonstrations and language instructions.
In the same vein as \cite{jang2022bcz}, we also use a multi-modal model for one-shot imitation, but we focus on zero-shot adaptation for different dynamics in a non-statinary environment. 
\section{Conclusion}
We presented a skill-based imitation learning framework, OnIS that enables not only one-shot imitation learning from a single demonstration but also zero-shot adaptation of a learned policy to different dynamics in a non-stationary environment. To this end, we leverage the compositionality of long-horizon tasks, by which a task can be decomposed into a sequence of  skills represented in the vision-language aligned semantic space, as well as explore meta-learning techniques to enable skill transfer upon various dynamics conditions. Through experiments, we demonstrated the superiority of OnIS with several one-shot imitation use cases in terms of one-shot imitation performance, generalization ability to unseen tasks and dynamics, and extensibility to different modalities in demonstrations. 
Our future direction includes addressing the issue of learning complex tasks via imitation in the environment with limited data, where sufficient expert data does not exist to learn diverse skills.

\section{Acknowledgement}
We would like to thank anonymous reviewers for their valuable comments and suggestions.
This work was supported by Institute of Information \& communications Technology Planning \& Evaluation (IITP) grant funded by the Korea government (MSIT)  
(No. 
2022-0-01045, 
2022-0-00043,  
2019-0-00421,  
2020-0-01821)  
and by the National Research Foundation of Korea (NRF) grant funded by the MSIT 
(No. RS-2023-00213118).  

\bibliographystyle{icml2023}
\bibliography{main}

\newpage
\appendix
\twocolumn



\section{Environments and Dataset}
\subsection{Multi-Stage Meta-world}
\textbf{Multi-stage Meta-world} is a robotic arm manipulation environment specifically designed for the execution of multi-stage tasks. 
For each task, multiple objects are positioned on a desk, where each object corresponds to existing tasks in Meta-world~\cite{yu2020metaworld}. The successful completion of a task is determined by the execution of $N$ subtasks in a predetermined sequence, similar to the Franka Kitchen environment~\cite{lynch2019playkitchen}.

\begin{figure}[h]
    \centering
    \includegraphics[width=0.9\columnwidth]{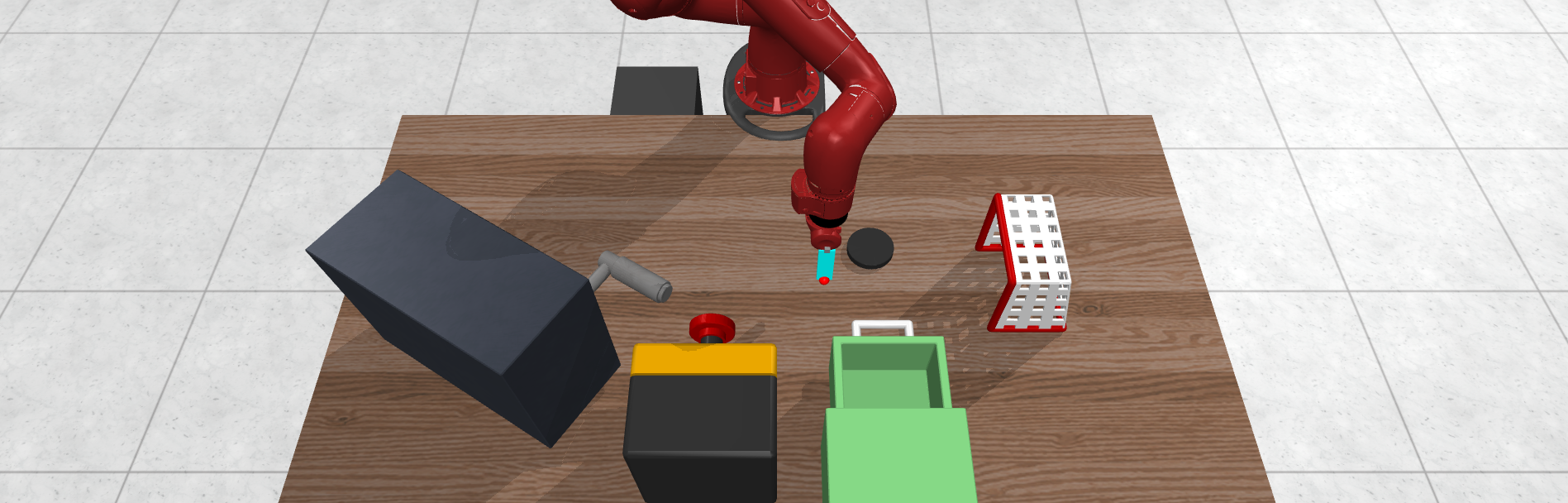}
    \caption{Multi-stage Meta-world}
    \label{fig:msmwMain}
\end{figure}

The state space is 140-dimensional, consisting of 4-dimensional position information of the robot arm, and 136-dimensional position (e.g., the location of the drawer) and status (e.g., whether the drawer is opened or closed) of objects. The action space is 4-dimensional, consisting of a 3-dimensional directional vector applied to the end-effector of the robot arm, and an 1-dimensional torque vector applied to the gripper. An example including the robot arm and several objects is illustrated in Figure~\ref{fig:msmwMain}. The reward function in the multi-stage Meta-world environment follows the same success metrics of Meta-world in a subtask level.

\textbf{Non-stationarity.} 
To emulate a non-stationary environment, we adopt a time-varying kinematic parameter of the robot arm.
Specifically, the environment involves such a kinematic parameter $w_t$ that varies 
for each timestep $t$, i.e.,
\begin{equation}
    \begin{split}
        & w_t = m + 0.25b \times sin(\rho_t)     \\
        & \rho_t = w_{t-1} + 0.75\pi \times z
    \end{split}
    \label{eqn:nonstationarity}
\end{equation}
where $w_0 = l$, and $z$ is a standard normal distribution $\mathcal{N}(0, 1)$. Note that $b$, $l$ and $m$ are hyperparameters to represent various configurations of environment dynamics.
Accordingly, given action $a_t$, the environment yields the next state which is sampled from the transition probability $s_{t+1}' \sim \mathcal{P}(\cdot | s_t, a_t + w_t)$.

\subsection{Expert Data Collection}
For expert data collection, we devise rule-based policies and execute them in a stationary environment. 
Regarding environment hidden dynamics, we use 13 different stationary environment configurations with dynamics parameters $b=0$ and $m \in \{-0.3, -0.25, -0.2, -0.15, -0.1, -0.05, 0, 0.05, 0.1, 0.15, \\ 0.2, 0.25, 0.3 \}$ in~\eqref{eqn:nonstationarity}.
Based on the initial location of objects and subtask execution orders, we have 24 different tasks. We use 16 tasks for training and use 8 tasks for evaluation, where for each task, 15 episodes are generated. 
Accordingly, we build a dataset of 15 episodes for each of the 16 tasks in 13 different environment settings with distinct dynamics parameters, containing 3,120 trajectories (by $16 \times 15 \times 13$).


\section{Experiment Details}
\subsection{OnIS Implementation}
\begin{table}[h]
    \caption{Hyperparameters for S-OnIS}
    \vskip 0.1in
    \label{hyperparameter:S-OnIS}
        \resizebox{\columnwidth}{!}
        {
            \begin{tabular}{c|c}
    	           \hline
                \textbf{Hyperparameter} & \textbf{Value} \\
                \hline
                Batch size & 1024 \\
                Prompt $\theta_p$ & 16 vector in 512-dim \\
                Frames $H$ of video clip & 30 \\

                Contrastive $\alpha$ (for \eqref{loss:prompt}) & 1 \\

                Dynamics encoder $\psi_{enc}$ & 3 FC with LSTM \\
                Activation function for $\psi_{enc}$ & Leaky ReLU \\
                Inverse dynamics decoder $\psi_{dec}$ & 5 FC with 128 units \\
                Activation function for $\psi_{dec}$ & Leaky ReLU \\
                The number of codebooks $M$ for $\psi_{enc}$ & 20 \\
                Dimension of codebook & 10 \\
                Learning rate for $\psi_{enc}$ & 5.55e-4 \\
                Length of sub-trajectory $H_0$ & 3 \\
                
                Semantic skill decoder $\Phi_{dec}$ & 4 FC with 128 units \\
                Activation function for $\Phi_{dec}$ & Leaky ReLU \\
                Learning rate for $\Phi_{dec}$ & 1e-4 \\
                Learning rate for $\psi_{dec}$ & 1e-4 \\

                Skill transfer module $\pi_{tr}$ & 5 FC with 128 units \\
                Activation function for $\pi_{tr}$ & Leaky ReLU \\
                Learning rate for $\pi_{tr}$ & 1e-4 \\
                \hline                
    	\end{tabular}
        }
\end{table}

\textbf{S-OnIS.}
S-OnIS involves 4-phases of training: 
(1) semantic skill sequence encoder $\Phi_{enc}$,
(2) semantic skill decoder $\Phi_{dec}$,
(3) dynamics encoder $\psi_{enc}$, and  
(4) skill transfer module $\pi_{tr}$.

During the training process of the semantic skill sequence encoder, the video demonstration being passed through the encoder $\Phi_{enc}$ creates a physical time bottleneck in the end-to-end training procedure of S-OnIS. To address this issue, we assign the inferred skill $z_t = \Phi_{enc}(v_{t: t+H})$ to each tuple $(s_t, v_t, l_t, a_t)$ in the dataset after training $\Phi_{enc}$. 

During the training process of the semantic skill decoder $\Phi_{dec}$, we have observed that the semantic skills within our dataset, consisting of long-horizon episodes, change infrequently (i.e., a maximum of 4 times).
As a result, the semantic skill decoder should predominantly predict 0 for most timesteps, leading to significantly lower performance in binary classification. To address this issue, we assign the label \textit{skill done} to each timestep in the dataset. This label indicates the end of a skill for a specific timestep as well as the preceding and subsequent 5-step intervals of the skill. Then, we have
\begin{equation}
    \textit{skill done}_t = 
    \begin{cases}
        1  & \text{if} \,\,\,\Phi_{enc}(v_{i: i+H}) \neq \Phi_{enc}(v_{i+1: i+1+H}) \\
        0  & o.w
    \end{cases}
\end{equation}
where $i \in [t-5, t+5]$.
The semantic skill decoder $\Phi_{dec}$ is trained to minimize binary cross entropy
\begin{equation}
    \mathcal{L}_{tp}(\tau) = \text{BCE}(\textit{skill done}_t, \Phi_{dec}(s_{t_0}, s_t, z_t))
\end{equation}
in~\eqref{termination:bce}. 

In the training procedure for the dynamics encoder $\psi_{enc}$, 
to enable backpropagation through non-differentiable quantization operator $\mathbf{q}$, we use the straight-through gradient estimator as explained in~\cite{NIPS2017vqvae}.
The skill transfer module $\pi_{tr}$ is implemented in a simple MLP which is trained via behavior cloning.
The hyperparameter settings for S-OnIS are summarized in Table ~\ref{hyperparameter:S-OnIS}.

\begin{table}[h]
    \caption{Hyperparameters for U-OnIS}
    \vskip 0.1in
    \label{table:U-OnIS_hyperparameter}
        \resizebox{\columnwidth}{!}
        {
            \begin{tabular}{c|c}
    	       \hline
                \textbf{Hyperparameter} & \textbf{Value} \\
                \hline
                Batch size & 1024 \\
                Encoder $\theta_v$ & Causal transformer \\
                Encoder $\theta_l$ & Causal transformer \\
                The number of layers for $\theta_v$, $\theta_l$ & 1 \\
                The number of heads for $\theta_v$, $\theta_l$ & 4 \\
                The number of codebooks $K$ for $\theta_l$ & 20 \\
                Learning rate for $\theta_v$, $\theta_l$ & 5e-5 \\
                Dimension of codebook for $\theta_l$ & 10 \\
                Length of sub-trajectory $H$ & 20 \\
                Contrastive $\alpha$ (for \eqref{loss:unsup}) & 1 \\
                Dynamics encoder $\psi_{enc}$ & 3 FC with LSTM \\
                Activation function for $\psi_{enc}$ & Leaky ReLU \\
                Inverse dynamics decoder $\psi_{dec}$ & 5 FC with 128 units \\
                Activation function for $\psi_{dec}$ & Leaky ReLU \\
                The number of codebook for $\psi_{enc}$ & 20 \\
                Dimension of codebook for $\psi_{enc}$ & 10 \\
                Learning rate for $\psi_{enc}$ & 5.55e-4 \\
                Learning rate for $\psi_{dec}$ & 1e-4 \\
                Length of sub-trajectory $H_0$ & 3 \\
                
                Semantic skill decoder $\Phi_{dec}$ & 4 FC with 128 units \\
                Activation function for $\Phi_{dec}$ & Leaky ReLU \\
                Learning rate for $\Phi_{dec}$ & 1e-4 \\

                Skill transfer module $\pi_{tr}$ & 5 FC with 128 units \\
                Activation function for $\pi_{tr}$ & Leaky ReLU \\
                Learning rate for $\pi_{tr}$ & 1e-4 \\
                \hline                
    	\end{tabular}
        }
\end{table}


\textbf{U-OnIS.}
U-OnIS involves 4-phases of training: 
(1) semantic skill sequence encoder $\Phi_{enc}$,
(2) semantic skill decoder $\Phi_{dec}$,
(3) dynamics encoder $\psi_{enc}$, and  
(4) skill transfer module $\pi_{tr}$.

In the training procedure for the semantic skill sequence encoder $\Phi_{enc}$, we use two additional encoders $\theta_v$ and $\theta_l$ which are similar to the skill predictor used in~\cite{lisa}, as explained in Section~\ref{section:semantic_skills_sequence}.
Encoder $\theta_l$ takes $(\Phi_l(l), s_{t-H: t})$ as input, and maps to a quantized semantic skill $z_t$ as
\begin{equation}
    \theta_l = \mathbf{q} \circ \theta^c_l: (\Phi_l(l), s_{t-H: t}) \mapsto z_t
\end{equation}
where $\theta^c_l$ maps the input $(\Phi_l(l), s_{t-H: t})$ to a continuous latent vector.
As explained in~\eqref{eqn:vector_quantization}, the quantization operator $\mathbf{q}$ 
maps the output of $\theta^c_l$ to one of the vectors in the skill codebook $\{ c_1, \cdots, c_K \}$.
That is, quantized semantic skill embedding $z_t$ at each timestep $t$ is defined as
%
\begin{equation}
    z_t = \underset{j \in \{1, \cdots, K\}}{\textnormal{argmin}} \{ \| \theta^c_l(\Phi_l(l), s_{t-H: t}) - c_j \| \}.
\end{equation}
Note that the contrastive loss presented in~\eqref{loss:unsup} is only responsible for updating the encoder $\theta_v$, and there is no gradient from~\eqref{loss:unsup} to update the encoder $\theta_l$.
The whole procedure of U-OnIS is presented in Algorithm~\ref{alg:deployment_uonis}  and the the hyperparameter settings for U-OnIS are summarized in Table~\ref{table:U-OnIS_hyperparameter}.

\begin{algorithm}[h]
    \caption{One-shot imitation of U-OnIS}
    \begin{algorithmic}[1]
    \STATE Semantic skill sequence encoder $\Phi_{enc}$
    \STATE Semantic skill decoder $\Phi_{dec}$
    \STATE Skill transfer module $\pi_{tr}$, Dynamics encoder $\psi_{enc}$
    \STATE Environment $env$, Demonstration $d$
    \STATE done $\gets$ False$,\ t \gets 0,\ \tau_t = queue(size=H_0)$
    \STATE $s_0 \gets env.\text{reset}()$
    \WHILE{not done}
        \STATE $z_t \gets \Phi_{enc}(d, s_{t-H: t})$ using \eqref{eqn:uonis_semantic_skill_encoder_forward}
        \STATE $h_t \gets \psi_{enc}(\tau_{t-1})$ using \eqref{eqn:quantization}
        \STATE $a_t \gets \pi_{tr}(s_t, h_t, z_t)$ using \eqref{eqn:skill_transfer}
        \STATE $s_{t+1}, \text{done} \gets env.\text{step}(a_t)$
        \IF {$\Phi_{dec}(s_{start}, s_t, z_t) = 1$}
            \STATE $s_{start} \gets s_t$
        \ENDIF
        \STATE $\tau_t.enqueue((s_t, a_t)),\ t \gets t + 1$
    \ENDWHILE
	\end{algorithmic}
	\label{alg:deployment_uonis}
\end{algorithm}



\subsection{Baseline Implementation}
\textbf{BC-Z.} 
BC-Z is an imitation learning framework which exploits multi-modal demonstrations.
In BC-Z, a language regression network is contrastively trained on an expert demonstration by maximizing the similarity between the embeddings of video demonstration and the embeddings of language instruction. 
Instead of training a video encoder, we use the embedding vectors extracted from the vision-language pertained model CLIP.
For a fair comparison, a history of state-actions is combined with either the vision or the language embedding as input to the policy.
%
%
The policy network is trained via behavior cloning, conditioned on video or language embeddings. 
The hyperparameter settings for BC-Z are summarized in Table~\ref{table:BCZ_hyperparameter}.%
\begin{table}[h]
    \caption{Hyperparameters for BC-Z}
    \vskip 0.1in
    \centering
    \label{table:BCZ_hyperparameter}
    \resizebox{\columnwidth}{!}
    {
        \begin{tabular}{c|c}
        \hline
        \textbf{Hyperparameter} & \textbf{Value} \\
        \hline
        Batch size & 1024 \\
        Language regression network & 5 FC with 512 units \\
        Length of expert sub-trajectory & 10 \\
        Learning rate & 1e-4 \\              
        \hline                
    	\end{tabular}
    }
    \end{table}
\textbf{Decision Transformer (DT).}
DT is a Transformer-based behavior cloning model.
Specifically, each token in language instruction is embedded into the latent vector using the pertained CLIP, and then it is concatenated with the state vector for the input of DT. The hyperparameter settings for DT are summarized in Table~\ref{table:DT_hyperparameter}.

\begin{table}[h]
\caption{Hyperparameters for DT}
\vskip 0.1in
\centering
\label{table:DT_hyperparameter}
    \begin{tabular}{c|c}
    \hline
    \textbf{Hyperparameter} & \textbf{Value} \\
    \hline
    Batch size & 64 \\
    The number of layers & 3 \\
    Embedding dimension & 128 \\
    The number of heads & 4 \\
    Dropout & 0.1 \\
    Learning rate & 1e-4 \\              
    \hline                
    \end{tabular}
\end{table}
\textbf{SPiRL.} SPiRL is an unsupervised skill learning framework, which embeds an expert sub-trajectory into a latent vector skill and decodes it into a sequence of actions via a primitive policy.
Specifically, the CLIP embedding vector of video is concatenated with a state and used as input of V-SPiRL, and the CLIP embedding of language instruction is concatenated with a state and used as input to L-SPiRL.
The hyperparameter settings for SPiRL are summarized in Table~\ref{table:SPiRL_hyperparameter}.
\begin{table}[h]
    \caption{Hyperparameters for SPiRL}
    \vskip 0.1in
    \centering
    \label{table:SPiRL_hyperparameter}
    \resizebox{\columnwidth}{!}
    {
        \begin{tabular}{c|c}
        \hline
        \textbf{Hyperparameter} & \textbf{Value} \\
        \hline
        Batch size & 1024 \\
        Skill encoder & 1 FC with LSTM \\
        Primitive policy & 7 FC with 128 units \\
        Length of expert sub-trajectory & 10 \\
        Skill dimension & 20 \\
        Learning rate & 1e-4 \\              
        \hline                
    	\end{tabular}
    }
\end{table}

\begin{figure*}[t]
    \centering
\includegraphics[width=0.9\textwidth]{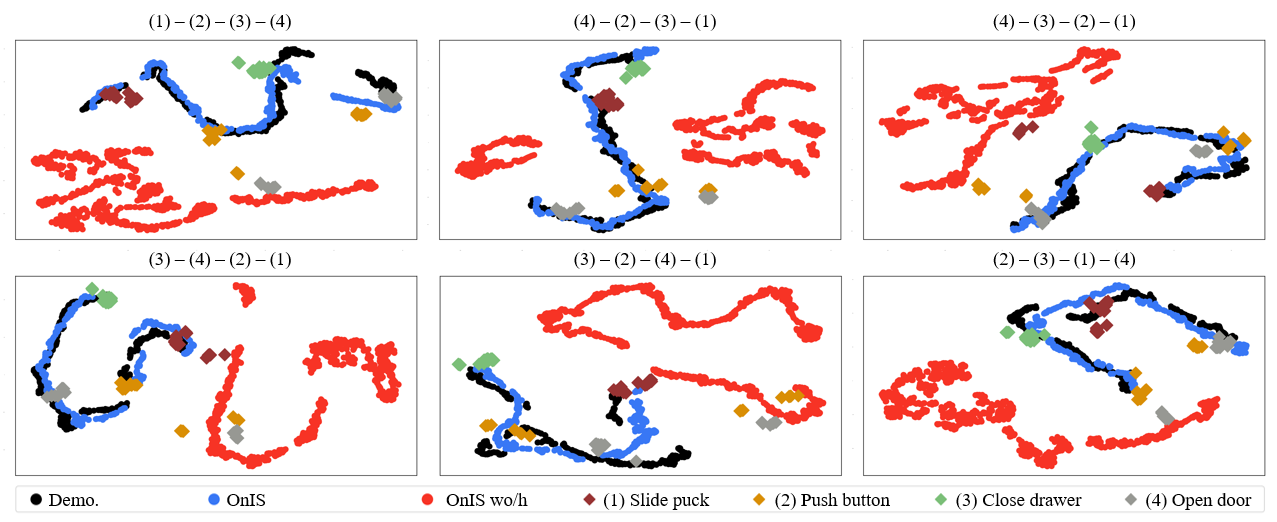}
    \caption{T-SNE embedding maps: (i) a video demonstration performing 4 subtasks (Demo.), (ii) a video generated by OnIS, and (iii) a video generated by OnIS without dynamics embedding (OnIS wo/h). Expert videos for each subtask (e.g., Slide puck, Push button, Close drawer, Open door) are represented in the same space.}
    \label{fig:tsne-feat-appendix}
\end{figure*}

\subsection{Evaluation Details}

The distribution of dynamics at test time assumes to be completely out of the distribution of training time dynamics. Specifically, the parameter $m$ is sparsely sampled (from 13 distinct values) according to uniform distribution of $[-0.3, 0.3]$ and $\omega_t$ is fixed in each trajectory at training time with $b=0$ in~\eqref{eqn:nonstationarity}. 
At test time, different hyperparameter values for $b, m$ are sampled at each episode from $[1, 2]$ and $[-0.45, 0.45]$, respectively. Accordingly, $\omega_t$ is changed by~\eqref{eqn:nonstationarity} at each timestep.

\subsection{Use Case Experiment Details}
\textbf{Noisy video demonstrations.}
We add noise to video demonstrations through three different methods (Color Jitter, Gaussian Noise, and Cutout) as described in~\cite{buslaev2020albumentations}.
\begin{itemize}
    \item \textbf{Color jitter.} In each video frame, adjustments are made to the brightness, contrast, saturation, and hue, in a way that random noise vectors are sampled from uniform distributions of $[-0.1, 0.1]$, $[-0.1, 0.1]$, $[-0.1, 0.1]$, and $[-0.025, 0.025]$ and the noise vectors are then added to video demonstrations.
    \item \textbf{Gaussian noise.} In each video frame, we sample a noise vector from a Gaussian distribution with a mean of 1 and a variance ranging from 1 to 10. The sampled noise is then added to video demonstrations.
    \item \textbf{Cutout.} 16 square regions of size $4 \times 4$ are randomly cropped in each video frame.
\end{itemize}
\textbf{Real-world video demonstration.}
%
%
We record various real-world video demonstrations in $700\times700$ resolution, which involve interactions with objects that represent different stages in multi-stage Meta-world. Then, we evaluate our framework with different noisy levels.
We also conduct the performance evaluation after fine-tuning the semantic skill sequence encoder (in the row of w/ Fine-tune in Table~\ref{tbl: realworld_noisy_finetune}), presuming the cases where fine-tuning is feasible if a real-world video demonstration is given with subtask-level instructions. 

\begin{table}[h]
    \centering
    \caption{Performance on real-world demonstrations}
    \label{tbl: realworld_noisy_finetune}
    \vskip 0.1in
    \resizebox{\columnwidth}{!}
    {
    \begin{tabular}{c|c|c|c}
    \hline
         & wo/ noise & noisy & very noisy \\
        \hline
        Success Rate & 91.67\% & 73.75\% & 54.38\%  \\
        Skills Matching Ratio & 100.0\% & 72.08\% & 52.92\%  \\
        \hline
        Success Rate (w/ Fine-tune) & 91.67\% & 83.13\% & 67.50\%  \\
        Skills Matching Ratio (w/ Fine-tune) & 100.0\% & 89.16\% & 76.46\%  \\
        \hline
    \end{tabular}
    }
    \vskip -0.1in
\end{table}
As observed in Table~\ref{tbl: realworld_noisy_finetune}, the skills matching ratio is positively correlated with the success rate, specifying that if correct semantic skills are predicted from the semantic skill sequence encoder, no performance degradation is expected for noisy real-world settings. When a single real-world demonstration is given with a subtask instruction, it is feasible to improve the success rate by tuning the language prompt of the semantic skill sequence encoder. As we test several real-world demonstrations with robotic manipulation in multi-stage Meta-world scenarios, we observe that the performance in the success rate can be rather dependent on the capability of underlying pretrained vision-language models. 

\textbf{Instruction variation.} 
Table~\ref{table:example_instruction_variation} describes the examples of the instruction variations in Table~\ref{table:novellanguage}.

\begin{table}[h]
	\label{table:languageExample}
	\caption{Examples of instruction variations}
    \vskip 0.1in
	\centering
        \resizebox{\columnwidth}{!}
        {
        \begin{tabular}{c}
    	        \hline
                Verb replacement \\
                \hline \hline
                push button $\rightarrow$ press the button \\
                open door $\rightarrow$ pull the door \\
                close drawer $\rightarrow$ shut the drawer \\
                slide puck $\rightarrow$ move the puck \\
                \hline
                Modifier extension \\
                \hline \hline
                push button $\rightarrow$ push the red button \\
                open door $\rightarrow$ open the black door \\
                close drawer $\rightarrow$ close the green drawer \\
                slide puck $\rightarrow$ slide the black puck \\
                \hline
                Verbose instruction \\
                \hline \hline
                push button $\rightarrow$ press the red button on yellow box \\
                open door $\rightarrow$ open the closed black door on the desk \\
                close drawer $\rightarrow$ close the green drawer on the desk \\
                slide puck $\rightarrow$ move the black plate to the white goal on the left side \\
                \hline
    	\end{tabular}
        }
\label{table:example_instruction_variation}
\end{table}

\section{Additional Experiments}
\subsection{Additional Experiments for Unseen Environments}
\begin{figure}[h]
    \centering
    \includegraphics[width=0.48\textwidth]{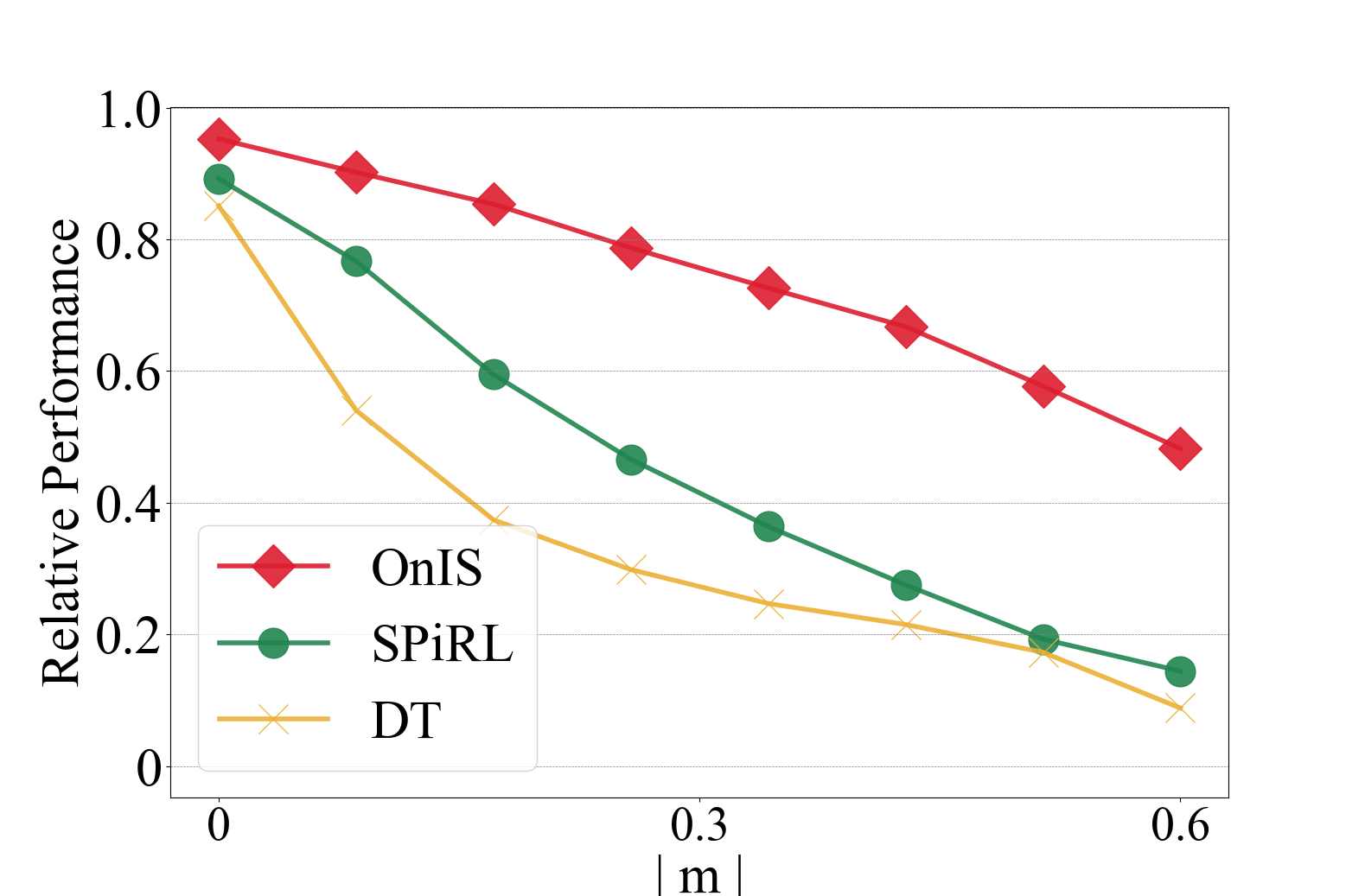}
    \vskip -0.1 in
    \caption{Relative performance of unseen dynamics to previously trained dynamics}
    \label{fig:dynamics_robustness}
\end{figure}

Table~\ref{table:language_stationary} shows the performance of OnIS when dynamics parameter $m$ in the seen task increases from $-0.6$ to $0$.
As shown, S-OnIS yields the best performance for all $K$ (the number of subtasks), implying that S-OnIS is robust to unseen dynamics.
This result corresponds to Section~\ref{subsec:oneshot_imitation_performance}.
Furthermore, Figure~\ref{fig:dynamics_robustness} shows that OnIS achieves the most robust performance as the dynamics level $|m|$ in~\eqref{eqn:nonstationarity} increases (e.g., to unseen dynamics) over previously trained dynamics.
This specifies that the dynamics embedding procedure of OnIS is able to disentangle dynamics information from expert trajectories.
\begin{table}[h]
    \caption{Performance in the success rate for multi-stage Meta-world with language instructions}
    \centering
    \vskip 0.1in
    \resizebox{\columnwidth}{!}
    {
        \begin{tabular}{c||c|c|c|c|c}
    	\hline
    	$K$ & L-DT & L-SPiRL & BC-Z & U-OnIS & S-OnIS \\
        \hline
        1 & 57.81\% & 29.68\% & 78.13\% & 70.32\% & \textbf{94.79\%} \\ 
        2 & 33.59\% & 37.50\% & 26.57\% & 66.57\% & \textbf{77.04\%} \\
        4 & 21.09\% & 17.96\% & 18.60\% & 47.44\% & \textbf{60.58\%} \\
        \hline
    	\end{tabular}
    }
	\label{table:language_stationary}
\end{table}
%

%

\subsection{Long-Horizon Multi-Stage Tasks}
\begin{table*}[t]
    \footnotesize
    \caption{Experiments for Long-Horizon Tasks}
    \label{tbl: long_horizon}
    \centering
    {
        \begin{tabular}{c | c|c|c|c|c|c|c|c|c|c}
        \hline
         & \multicolumn{10}{c}{Number of objects ($K$)}  \\ 
         \cline{2-11}
         & 1 & 2 & 3 & 4 & 5 & 6 & 7 & 8 & 9 & 10 \\
         \hline
         OnIS(Language Instruction) & 99.33\% & 76.0\% & 74.66\% & 69.33\% & 65.33\% & 58.0\% & 50.66\% & 44.0\% & 36.0\% & 26.0\% \\
         \cline{1-1}
         OnIS(Video Demonstration) & 99.33\% & 72.0\% & 64.0\% & 60.33\% & 42.0\% & 29.0\% & 10.03\% & 4.0\% & 1.33\% & 0.0\% \\
         \hline
        \end{tabular}
    }
        \vskip -0.1in
\end{table*}

To investigate whether our approach can tackle long-horizon multi-stage tasks with a large number of skills, we evaluate OnIS with manipulation tasks that involve up to 10 objects. As shown in Table~\ref{tbl: long_horizon}, OnIS is able to solve the tasks with up to 10 objects ($K=1,\dots,10$). This clarifies that the semantic skill sequence encoder scales well to  predict semantic skills even when a task involves multiple subtasks of up to 10.

\subsection{Visualization of Semantic Reconstruction}
Figure~\ref{fig:tsne-feat-appendix} represents the embedding maps of the expert demonstration (Demo.), U-OnIS (OnIS), and OnIS trained without dynamics embedding (OnIS wo/h) for 6 different tasks.
Similar to the experiment in Section~\ref{expl:semantic_embedding_correspondence} in the main manuscript, the embeddings of episodic run by U-OnIS and demonstrations 
show highly similar paths, meaning that their semantic skills are aligned. On the other hand, the OnIS without dynamics embedding has low  semantic skill correspondence with demonstrations.

\subsection{Additional Baseline: Multigame-DT}
\begin{table}[h]
    \caption
    {
    One-shot imitation performance for language instruction 
    }
    \vskip 0.1in
    \footnotesize
    \centering
    \resizebox{\columnwidth}{!}
    {
        \begin{tabular}{c|c||c|c|c||c|c}
        \hline

        \multicolumn{1}{c|}{$K$}  & \multicolumn{1}{c||}{Non-st.} 
        & \multicolumn{1}{c|}{Multi-DT} & \multicolumn{1}{c|}{L-DT}
        & \multicolumn{1}{c||}{BC-Z} & \multicolumn{1}{c|}{U-OnIS} & \multicolumn{1}{c}{S-OnIS} \\
        \hline

        \multirow{4}{*}{1} &
        Stationary      & 100.0\% & 92.66\% & 100.0\% & 87.70\% & 100.0\% \\ \cline{2-7}
        & Low           & 85.58\% & 62.49\% & \textbf{96.61\%} & 90.40\% & 95.00\% \\  
        & Medium        & 73.07\% & 53.84\% & 75.32\% & 81.29\% & \textbf{89.04\%} \\  
        & High          & 66.34\% & 20.19\% & 68.91\% & 82.19\% & \textbf{88.50\%} \\  
        \hline

        \multirow{4}{*}{2} & 
        Stationary      & 68.75\% & 58.33\% & 76.25\% & 74.50\% & \textbf{95.00\%} \\ \cline{2-7}
        & Low           & 53.84\% & 47.12\% & 45.27\% & 60.44\% & \textbf{83.17\%} \\
        & Medium        & 52.40\% & 33.17\% & 35.00\% & 69.24\% & \textbf{78.50\%} \\  
        & High          & 42.79\% & 30.28\% & 25.52\% & 66.67\% & \textbf{77.27\%} \\ 
        \hline

        \multirow{4}{*}{4} & 
        Stationary      & 31.25\% & 37.73\% & 40.30\% & 67.71\% & \textbf{87.50\%} \\ \cline{2-7}
        & Low           & 20.55\% & 22.11\% & 20.23\% & 59.64\% & \textbf{76.60\%} \\
        & Medium        & 18.51\% & 20.43\% & 20.60\% & 56.68\% & \textbf{71.39\%} \\
        & High          & 14.90\% & 8.90\% & 11.13\%  & 54.82\%  & \textbf{66.92\%} \\  
        \hline
        \end{tabular}
    }
    \label{tbl:multi_dt}
        \vskip -0.1in
\end{table}

We perform the experiment with the latest version of DT, a multigame-DT ~\cite{multigame_dt}. Table~\ref{tbl:multi_dt} shows the performance in multi-stage Meta-world achieved by OnIS and baselines including Multi-DT. 
We observe that Multi-DT achieves better than L-DT with an average gain of 11.73\%, but shows worse performance than S-OnIS with an average of loss of 31.74\%
For fair comparisons, we use video and language embeddings from the pretrained CLIP model as an additional input to Multi-DT, as well as other baselines. Since Multi-DT uses rewards in offline datasets (unlike OnIS without any rewards), we consider that Multi-DT has some advantage over the original DT in terms of policy learning. Otherwise, Multi-DT becomes similar to the original DT. In our problem setting of one-shot imitation for multi-stage tasks, Multi-DT is yet limited in exploring task compositionality. 

\subsection{Ablation Studies}

\begin{table*}[t]
    \caption{Ablation: Contrastive Learning on Dynamics Encoder}
    \centering
    \label{tbl: ablation_contrastive}
    {
        \begin{tabular}{c|c||c|c||c|c || c}
        \hline
        \multicolumn{1}{c|}{$K$}  
        & \multicolumn{1}{c||}{Non-st.} 
        & \multicolumn{1}{c|}{w/ Contra, w/ VQ} 
        & \multicolumn{1}{c||}{w/ Contra, wo/ VQ}
        & \multicolumn{1}{c|}{wo/ Contra, w/ VQ}
        & \multicolumn{1}{c||}{wo/ Contra, wo/ VQ}
        & \multicolumn{1}{c}{wo/ Recon}
        \\
        \hline

        \multirow{4}{*}{1} &
        Stationary      & 100.0\% & 100.0\% & 100.0\% & 100.0\% & 100.0\% \\ \cline{2-7}
        & Low           & \textbf{95.01\%} & 93.84\% & 92.30\% & 89.42\% & 89.10\% \\  
        & Medium        & \textbf{94.55\%} & 92.11\% & 83.65\% & 77.24\% & 84.40\% \\  
        & High          & 88.50\% & \textbf{89.42\%} & 72.62\% & 61.21\% & 81.76\%\\  
        \hline

        \multirow{4}{*}{2} & 
        Stationary      & 100.0\% & 100.0\% & 68.75\% & 79.16\% & 95.10\% \\ \cline{2-7}
        & Low           & \textbf{85.82\%} & 83.75\% & 66.98\% & 76.60\% & 75.04\% \\
        & Medium        & \textbf{84.42\%} & 82.37\% & 61.37\% & 58.52\% & 74.45\% \\  
        & High          & \textbf{81.14\%} & 69.07\% & 49.34\% & 48.96\% & 75.04\% \\ 
        \hline

        \multirow{4}{*}{4} & 
        Stationary      & 87.50\% & \textbf{88.09\%} & 46.88\% & 58.34\% & 85.50\% \\ \cline{2-7}
        & Low           & \textbf{80.11\%} & 74.45\% & 45.97\% & 53.50\% & 71.96\% \\
        & Medium        & 75.24\% & \textbf{75.58\%} & 43.50\% & 44.31\% & 67.04\% \\
        & High          & \textbf{66.92\%} & 63.19\% & 34.39\% & 34.13\% & 65.32\% \\  
        \hline
        \end{tabular}
    }
    \vskip -0.1in
\end{table*}

\textbf{Contrastive learning on dynamics encoder.}
To investigate the effect of contrastive learning-based dynamics embedding in our proposed OnIS framework, we  evaluate and compare several different implementations for dynamics embedding as part of ablation. Table~\ref{tbl: ablation_contrastive} illustrates those including the cases of whether the contrastive loss is added to the typical dynamics embedding loss or not (w/ Contra and wo/ Contra, respectively, in the column names), the cases of whether the dynamics embedding vector $\psi_{enc}(\tau)$ is quantized or not (w/ VQ and wo/ VQ, respectively, in the column names), and the case of reconstruction loss \eqref{equ:L_REC} is not added (wo/ Recon, in the column name) to the dynamics embedding loss.

Here, the typical dynamics embedding loss is based on reconstruction, as in the formula in~\eqref{equ:L_REC}, similar to several related works such as~\cite{varibad, woo_nonstationary, xie_continual}:
\begin{equation}
    \mathcal{L}_{DE} = \mathbb{E}_{t-H_0 < i \leq t}\left[\|a_i - \psi_{dec}(s_i, s_{i+1}, \psi_{enc}(\tau_t)\|^2\right]
    \label{eqn:typical_de}
\end{equation}
where $\psi_{enc}$ and $\psi_{dec}$ denote the dynamics encoder and decoder, respectively. 

As observed in Table~\ref{tbl: ablation_contrastive}, the contrastively trained dynamics encoder (w/ Contra) improves the performance by 19.8\% in average success rates for multi-stage Meta-world tasks, compared to the dynamics encoder trained without contrastive learning (wo/ Contra). 

Moreover, we also observe that quantizing the output of the dynamics encoder (w/ VQ) improves the performance by 2.2\% in average success rates, compared to the dynamics embedding without quantization (wo/ VQ). Additionally, the reconstruction loss in the training of dynamics encoder (w/ Contra, w/VQ) improves the performance by 6.21\%, compared to the dynamics encoder trained without reconstruction loss (wo/ Recon).

These ablation results with such performance gains specify that contrastive learning enables the dynamics encoder to disentangle task-invariant dynamics-relevant features from expert demonstrations. The task information - what to do, such as open door and push button - is extracted by the semantic skill sequence encoder in prior, and then it is feasible to enforce different sub-trajectories sharing the same dynamics have a close proximity in the dynamics embedding space, regardless of which skills the sub-trajectories involve. 
As two sub-trajectories in a trajectory are used as a positive pair in our sampling strategy, the contrastive learning loss in~\eqref{equ:L_CON} renders the aforementioned desired property of the dynamics embedding space.


\end{document}